\documentclass[conference]{IEEEtran}
\IEEEoverridecommandlockouts

\usepackage{amsmath,amssymb,amsfonts}
\usepackage{algorithmic}
\usepackage{graphicx}
\usepackage{textcomp}
\usepackage{xcolor}
\usepackage{subcaption}
\usepackage{array}
\usepackage{adjustbox}
\usepackage{booktabs}
\usepackage{makecell}

\usepackage[style=ieee]{biblatex}
\begin{document}

\title{PCB-Vision: A Multiscene RGB-Hyperspectral Benchmark Dataset of Printed Circuit Boards}

\author{Elias~Arbash, Margret~Fuchs, Behnood~Rasti, Sandra~Lorenz, Pedram~Ghamisi, Richard~Gloaguen 
\thanks{Elias Arbash, Margret Fuchs, Behnood Rasti, Sandra Lorenz, Pedram Ghamisi, and Richard Gloaguen are with the Helmholtz-Zentrum Dresden-Rossendorf (HZDR), Helmholtz Institute Freiberg for Resource Technology (HIF), 09599 Freiberg, Germany.}%
}%

\maketitle

\begin{abstract}

Addressing the critical theme of recycling electronic waste (E-waste), this contribution is dedicated to developing advanced automated data processing pipelines as a basis for decision-making and process control. Aligning with the broader goals of the circular economy and the United Nations (UN) Sustainable Development Goals (SDG), our work leverages non-invasive analysis methods utilizing RGB and hyperspectral imaging data to provide both quantitative and qualitative insights into the E-waste stream composition for optimizing recycling efficiency.
In this paper, we introduce 'PCB-Vision'; a pioneering RGB-hyperspectral printed circuit board (PCB) benchmark dataset, comprising 53 RGB images of high spatial resolution paired with their corresponding high spectral resolution hyperspectral data cubes in the visible and near-infrared range. Grounded in open science principles, our dataset provides a comprehensive resource for researchers through high-quality ground truths, focusing on three primary PCB components: integrated circuits (IC), capacitors, and connectors. We provide extensive statistical investigations on the proposed dataset together with the performance of several benchmark state-of-the-art models (SOTA), including U-Net, Attention U-Net, Residual U-Net, LinkNet, and DeepLabv3+.
By openly sharing this multi-scene benchmark dataset along with the baseline codes, we hope to foster transparent, traceable, and comparable developments of advanced data processing across various scientific communities, including, but not limited to, computer vision and remote sensing. 
Emphasizing our commitment to supporting a collaborative and inclusive scientific community, all materials, including code, data, ground truth, and masks, will be accessible at 
\url{https://github.com/hifexplo/PCBVision}.

\end{abstract}

\begin{IEEEkeywords}
circular economy, automated data processing, optical sensors, recycling, e-waste, printed circuit board, hyperspectral, dataset, RGB, conveyor belt, sensors, machine learning, deep learning, PCBVision, open-source data, digitalization
\end{IEEEkeywords}

\begin{figure*}[t]
\centering
\includegraphics[width=\textwidth]{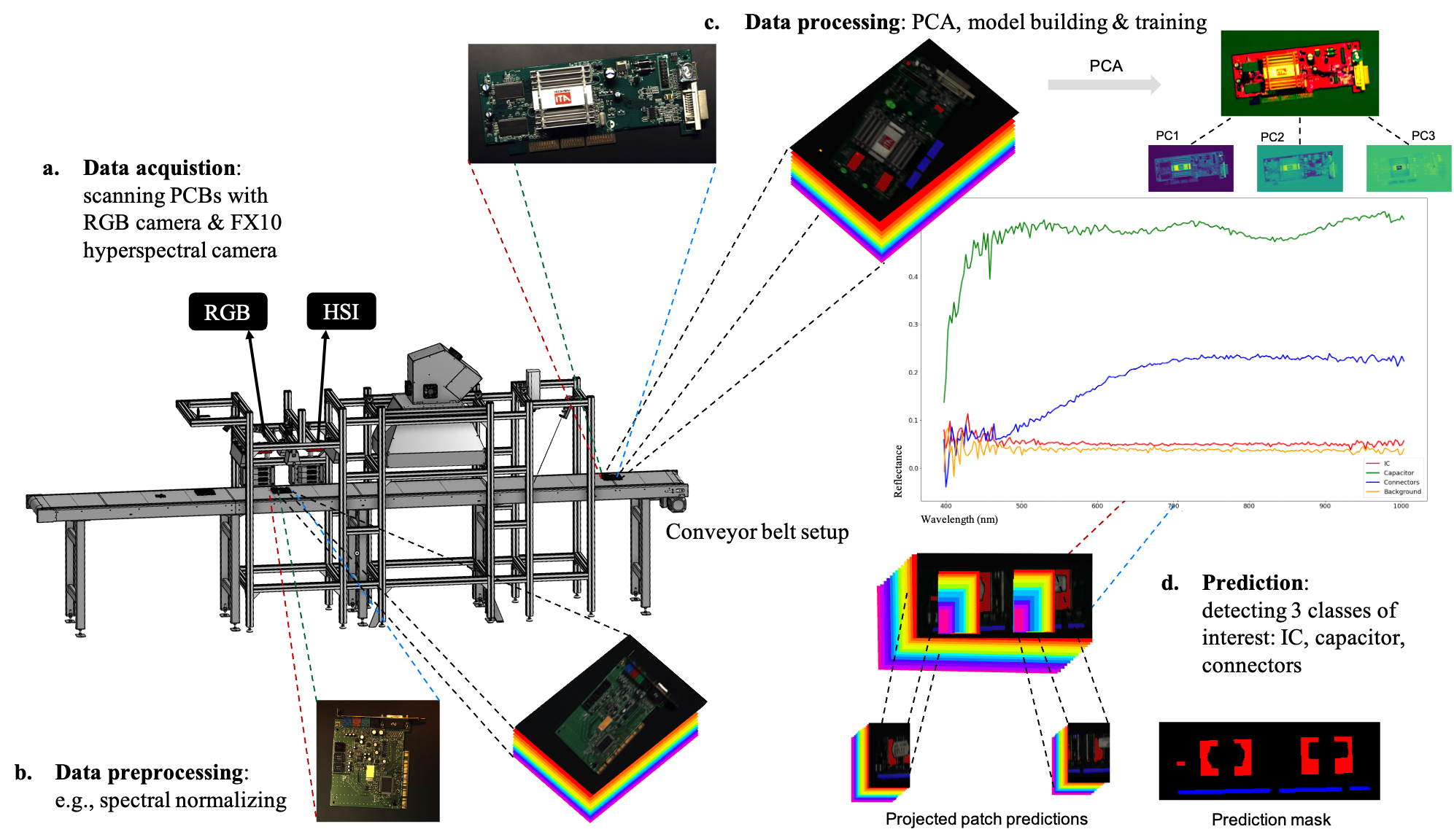}
\caption{PCB-Vision setup to results:  (a) RGB images and HS data cubes are acquired, (b) data normalization and preprocessing, (c) data preparation for ML model pipeline, (d) segmentation results}
\label{fig:workflow}
\end{figure*}

\section{\textbf{Introduction}}

The electronics market has witnessed remarkable growth and development over the last few years, driven by the high demand for new generations of electronic devices. However, this accelerated technological advancement has resulted in a considerably reduced lifespan for electronic products, leading to a surge in electronic waste (E-waste) \cite{zlamparet2017remanufacturing}. Research studies and journal reports indicate that the global E-waste generation has been escalating at an unprecedented rate, posing significant environmental challenges and sustainability concerns \cite{chinaCE}.
It is estimated that each year around 30 to 50 million tons of waste from electrical and electronic equipment (WEEE) are disposed, with an annual growth rate estimation of 3 to 5\% \cite{cucchiella2015recycling}. In 2019, a staggering 53.6 million metric tons of E-waste was generated globally, with only 17.4\% of it officially documented as recycled \cite{ewaste}.  This resulting accumulation of E-waste, along with its unrecovered critical and toxic raw materials demands effective and efficient recycling strategies to mitigate its environmental impact and harness the economic value hidden within E-waste components.

The concept of a circular economy provides a framework for addressing E-waste challenges by promoting the recovery, and reuse of materials \cite{Lieder2016Towards}. E-waste recycling specifically plays a crucial role in this transition by transforming discarded electronic devices into valuable resources.
E-waste recycling also contributes to the achievement of several sustainable development goals (SDGs) outlined by the United Nations (UN) \cite{undp2023}. 
E-waste recycling reduces primary resource consumption and waste generation at the same time corresponding to SDG 12 "Responsible Consumption and Production: Ensure Sustainable Consumption and Production Patterns". Moreover, it helps reduce greenhouse gas emissions associated with virgin material extraction and processing supporting SDG 13 "Climate Action". Additionally, promoting the reuse of existing materials to reduce the environmental impact of virgin resource extraction aligns with SDG 15 "Life on land". 

Among the myriad electronic components contributing to E-waste, printed circuit boards (PCBs) hold particular significance due to their widespread use in various electronic devices. The recycling of waste PCBs can unlock significant value, given their considerable residual content. Approximately 28\% of a PCB's weight is constituted by high-grade precious metals such as Au, Ag, Cu, Pd, and Ta \cite{chen2019recycling}. The extraction of those metals can help reduce the need for raw material extraction. Nevertheless, waste PCBs present environmental and health hazards. The methods employed for extracting precious metals, particularly in open-air settings like PCB acid baths \cite{Guo2009Recycling}, can inadvertently release toxic substances, including lead and mercury into the environment \cite{chen2019recycling}.

To achieve efficient and sustainable recycling of PCB components, i.e., an optimized E-waste recycling, there arises a pressing need for automated advanced analytical techniques and informative systems that can provide both qualitative and quantitative information in a short time about the PCB composition. Smart systems that adopt optical sensors, including but not limited to RGB cameras, have demonstrated remarkable potential in this domain \cite{MLPCB}. By leveraging the rich data and information obtained from these sensors, machine learning (ML) and deep learning (DL) methods are to be deployed to streamline the recycling process and improve its accuracy and efficiency \cite{DLPCB}.
Motivated by these technological and computational advancements, our project Ramses for Circular Economy (Ramses-4-CE) spearheads the development of optical spectroscopy-based multi-sensor systems tailored for the E-waste recycling industry. At the core of our initiative lies the pursuit of advanced multi-source data fusion e.g., RGB and hyperspectral imaging sensors. The usage of ML techniques for data fusion enables rapid data integration and automates information extraction from E-waste, thereby optimizing E-waste recycling practices.

RGB cameras offer several advantages, such as cost-effectiveness, high spatial resolution, low integration time, ease of use, and real-time data acquisition, making them popular choices for various PCB applications, such as defect detection and quality control. 
These cameras excel in capturing visual information at a sub-millimeter pixel scale and are widely utilized in image-based inspection systems within the electronics industry [\cite{pearson2009hardware}, \cite{Jahanshahi2009ASA}]. 

Work examples where RGB cameras contributed to PCBs-oriented informative systems are Herchenbach et al. \cite{herchenbach2013segmentation} who used the RGB and the depth information of the Microsoft Kinect sensor to segment and classify the through-hole components (THC) mounted on printed circuit board assembly (PCBA).
Li et al. \cite{li2013smd} proposed an information retrieval automated PCB recycling system by detecting and segmenting surface-mounted devices (SMDs) of two kinds, small devices like resistors and capacitors, and integrated circuits (IC). 
In another work, Li et al. \cite{li2014text} also enhanced the text information retrieval of optical character recognition (OCR) from RGB images by developing a novel thresholding method that utilizes an adaptive window size along with background estimation.  Li et al. work proves that text recognition quality can be improved by enhancing the binarization of text contents.
For PCB inspection and fault detection systems, Kim et al. \cite{defectdet} developed a skip-connected convolutional autoencoder that took RGB images as input to detect defects in PCB for automating PCB surface inspection. 
Ding et al. \cite{ding2019tdd} propose a tiny defect detection network (TDD-Net) utilizing deep convolutional neural networks (CNN) to enhance PCB quality control systems using RGB images. 
In \cite{adibhatla2020defect} Adibhatla et al. used a version of you-only-look-once (YOLO) \cite{redmon2016you} named tinyYOLOv2 DL algorithm for defect detection in PCBs using 11,000 RGB images for PCB quality inspection. 

\begin{table*}[t]
\caption{A systematic review of publicly available PCB datasets}
\begin{adjustbox}{width=0.98\textwidth}
    \centering
        \begin{tabular}{ccccccc}
         \toprule
             \textbf{Dataset}&  \textbf{PCB\#} &  \textbf{Data Volume}&  \textbf{Data Type}&  \textbf{Derivatives of Original} &  \textbf{Inspected Object}& \textbf{Sensor Type}\\
             \midrule
              FICS-PCB \cite{lu2020fics}&  31&  9,912&  RGB Images&  Illumination, scale, sensor&  \makecell{Capacitor, resistor, inductor, \\ transistor, diode, IC}  
& Digital microscope + DSLR
\\
\makecell{\\ FICS-PCB X-ray \cite{mehta2022fics}}
&  5&  5&  3D volumetric data
&  -&  PCBs& NANO-CT-GE V | TOME \ X M 240\\
             \makecell{\\ PCBA-defect \cite{huang2019pcb}}
&  -&  1,386&  RGB images&  -&  6 Trace defects& Digital microscope
\\
             \makecell{\\ Deep PCB \cite{tang2019online}}
&  -&  1,500&  Binary images&  -&  6 Trace defects
& Linear scan CCD
\\
             \makecell{\\ PCB-DSLR \cite{pramerdorfer2015dataset}}
&  165&  748&  RGB images&  Rotation&  PCB + (IC)& DSLR\\
             \makecell{\\ PCB-Metal \cite{Mahalingam2019PCBMETALAP}}
&  123&  984&  RGB images&  4 rotations + front and back scan&  \makecell{(IC, capacitors,\\ resistors, inductors)}
& DSLR\\
             \makecell{\\ PCB-Vision}&  53&  106&  RGB images + hyperspectral cubes&  Provided (optional)& \makecell{ PCB + (IC, \\ capacitor, connectors)}
& DSLR + Linescan Spectrometer\\
\bottomrule
        \end{tabular}
        \end{adjustbox}
    \label{table:datasets}
\end{table*}

In contrast to RGB, hyperspectral imaging (HSI) cameras bring an abundance of benefits for PCB component detection and inspection systems, e.g. offering comprehensive spectral coverage that enables the identification and analysis of materials based on their unique spectral signatures \cite{Willett2014Sparsity}. 
Several studies employed spectroscopy-based methods to analyze and map PCB compositions, e.g., Englert et al. in \cite{englert2021use} proposed a monitoring method to detect and quantify organic contaminations on technical surfaces using hyperspectral imaging and X-ray photoelectron spectroscopy (XPS). 
Carvalho et al. \cite{CARVALHO2015278} examined and analyzed 1,200 emission points on a 30mmx40mm section of a 2011 mobile phone PCB sample within the wavelength range of 186 to 1040 nm using laser-induced breakdown spectroscopy (LIBS) and scanning electron microscopy with energy-dispersive X-ray spectroscopy (SEM-EDS). Their investigation focused on determining the metal compositions, and a graphical map of the metal distribution was obtained. 
In a study done on end-of-life mobile phone wastes, Palmieri et al. \cite{palmieri2014recycling} proved that the characterization which combines traditional methods like scanning electron microscopy and Raman spectroscopy alongside innovative hyperspectral imaging in the short-wave infrared range significantly enhances recycling strategies and product recovery outcomes.
Rapolti et al.  \cite{rapolti2021experimental} developed a two-stage sorting stand for E-waste, employing optical sensors in both stages. The first stage utilizes a Specim near-infrared FX10e hyperspectral camera and a robotic arm. Unrecognized items proceed to the second stage, which employs an IFM Electronics contour vision sensor and actuators for further sorting.
Sudharshan et al. \cite{sudharshan2020object} proposed an enhanced version of the FasterRCNN object detection method called GOL-based Faster-RCNN that utilizes RGB and FX17 HSI camera information for enhancing the object detection performance of PCB components for aiding the recycling and recovery systems of PCB.
Polat et al. \cite{polat2021combined} fused 3D point clouds and HSI for classifying different types of objects and materials of electronic waste like shredded PCBs and plastics, this usage allowed the combination of geometrical and physical information to help components distinguish routines of PCB components. 

The aforementioned systems enrich the research towards building efficient non-invasive PCB informative systems that act as a cornerstone for efficient PCB recycling streams. However, a foundational element in the development of those advanced systems is having comprehensive datasets that act as the bedrock upon which algorithms and models are designed and tested. To our knowledge, only a few PCB datasets are publicly available e.g., the publicly available 'FICS-PCB' dataset developed by Mehta et al. \cite{lu2020fics} stands out as a valuable resource for the development of robust PCB automated visual inspection (AVI) systems. Containing 9,912 RGB images with 77,347 annotated components from 31 distinct PCBs, the FICS-PCB dataset provides a comprehensive platform for evaluating and improving the performance of PCB AVI algorithms. The authors demonstrated the effectiveness of two deep learning (DL) architectures, AlexNet \cite{krizhevsky2012imagenet}, and Inception-v3 \cite{szegedy2016rethinking}, on the FICS-PCB dataset, achieving promising results in PCB components classification.
Mehta et al. \cite{mehta2022fics} also proposed another dataset 'FICS-PCB X-ray' that supports the PCB-AVI which is the first annotated X-ray PCB dataset for automated PCB inter-layer inspection including 5 PCBs each containing 3D volumetric data that can be extracted into 2D slices for the analysis. 
In the field of PCB automated optical inspection (AOI), Huang et al. \cite{huang2019pcb} proposed a PCB dataset referred to as 'PCBA-Defect' for defect detection and classification containing 1,386 RGB images with six kinds of synthetic defects (missing hole, mouse bite, open circuit, short, spur and spurious copper) photoshopped on cropped images of scanned PCB traces using an industrial camera equipped with a CMOS sensor and a zoomable industrial lens.
In \cite{tang2019online} Tang et al. proposed 'DeepPCB', a dataset containing 1,500 binary image pairs of defect-free and defected traces with annotations of six common types of defects (open, short, mouse bite, spur, pinhole, and spurious copper) along their positions, moreover a DL model utilizing a novel group pyramid pooling was proposed to efficiently detect the PCB trace defects. 
To facilitate research on computer-vision-based PCB analysis, Pramerdorfer et al. \cite{pramerdorfer2015dataset} proposed a PCB dataset referred to as 'PCB-DSLR' containing 748 images of 165 different PCBs using a digital single-lens reflex (DSLR) camera scanned in an industrial-like environment on top of a conveyor belt. 'PCB-DSLR' is also provided with accurate PCB segmentation information, in addition to 9,313 bounding boxes for detecting integrated circuits (ICs), along with textual information of the labels of 1,740 IC samples. A similar to \cite{pramerdorfer2015dataset} but more abundant PCB dataset is Mahalingam et al. \cite{Mahalingam2019PCBMETALAP} who introduced 'PCB-Metal' (PCB Metal is not available to this publication date); a 984 high-resolution RGB images PCB dataset of 123 different PCBs with object detection ground truths of 4 main components of PCBs (IC, capacitors, resistors, and inductors) useful for image-based PCB analysis, e.g., PCB classification, component detection, etc. Table \ref{table:datasets} provides a summary of the publicly available PCB datasets with their sensor and data types information according to our knowledge.

It can be noted that, although reflectance-based HSI provides further features (spectral representation of a physical absorption process) that allow more advanced vital analysis for the PCB recycling-oriented system, the image processing progression of HSI lags behind its RGB counterpart due to the scarcity of available data. The advancement of models and methodologies in HSI requires large datasets, facilitating enhanced development of non-invasive PCB inspection systems. 

Existing HSI datasets remain sparse, many are confined to a single scene, and highly non-industrialized applications are rather more focused on food \cite{feng2012application}, medical fields \cite{calin2014hyperspectral}, and earth observation e.g., Indian Pines, Houston, Salinas scene, Kennedy Space Center, and Pavia University. Those datasets formed a baseline to develop many of the HSI classification modalities e.g., 'HybridSN' \cite{roy2019hybridsn} a model by Roy et al. that combines 3D CNN with 2D CNN for rich extraction spatial-spectral feature representations. The utilization of 3D CNN on the full HSI scene on a graphical processing unit (GPU) demands a lot of unavailable memory, therefore patching from the scene was fed to the model. For unshaped objects e.g., earth observation scenes, this is not a problem, but it might be for geometrically shaped objects like PCBs as we discuss later on.
Another example is Uchaev et al.  \cite{uchaev2023small} who presented RPNet-RF, a model  that combines recursive filtering and random patches network (RPNet) to extract informative features that will be combined with the HSI spectral features to classify the HSI using a support vector machine (SVM) classifier. 
Nonetheless, the lack of multiscene hyperspectral (HS) data hinders the development of HSI image processing models with high generalization ability to unseen scenes, therefore, multiscene hyperspectral datasets are needed. The multiscene HSI datasets allow models to handle more illumination variations, noise (ambient and instrument), and variable proportions of content (objects/ classes), which challenge existing models trained on single-scene datasets.
Recently, more sophisticated multiscene HSI datasets emerged, for instance, 'landslide4sense' presented by  Ghorbanzadeh et.al \cite{ghorbanzadeh2022landslide4sense} where 3,799 HSI patches of size 128 by 128 acquired by fusing optical layers from the Sentinel-2 sensor collected at four different times and geographical locations: Iburi (2018), Kodagu (2018), Gorkha (2015), and Taiwan (2009). Landslide4sense provides binary classification ground truth of landslide or non-landslide labels to facilitate accurate detection of landslide extents.
Further remote sensing urban environment multiscene dataset is 'C2Seg' provided by Hong et al. \cite{hong2023cross}. C2Seg is a multimodal multiscene benchmark dataset consisting of two cross-city scenes; Berlin-Augsburg (Germany) and Beijing-Wuhan (China) for the sake of cross-city semantic segmentation studies. C2Seg contains multiple data types including hyperspectral, multispectral, and synthetic aperture radar (SAR).
Such datasets support the development of high-in-generalization HS data classification models since they allow classifiers to be trained on data from multiple scenes, i.e., considering further variance in the targets which improves the prediction performance on unseen data.


Nevertheless, all the above-mentioned datasets and investigations do not provide a simultaneously comprehensive presentation of PCBs across both RGB and HSI domains. Addressing this gap, we introduce 'PCB-Vision', a multiscene multimodal (HSI and RGB) dataset consisting of 53 high spectral resolution hyperspectral data cubes along with their 53 high spatial resolution RGB images of 53 different PCBs scanned in industrial-like scenarios, coupled with two different segmentation annotations. In summary, PCB-Vision consists of:
\begin{enumerate}
    \item 53 hyperspectral cubes of 53 different PCBs scanned in the visible and near-infrared range with 224 bands.
    \item 53 RGB images of the 53 PCBs.
    \item  'General' pixel-wise classification ground truth of three classes of interest ('IC', 'Capacitor', 'Connectors') for both RGB and HS data cubes.
    \item 'Monoseg' pixel-wise classification ground truth of three classes of interest ('IC', 'Capacitor', 'Connectors') for both RGB and HS data cubes.
    \item 53 background/foreground masks for separating the PCB from the image background. 
\end{enumerate}

The integration of RGB and HS data within a single dataset holds significant promises, such datasets foster cross-modal exploration, empowering researchers and practitioners to forge novel pathways for material stream digitalization, monitoring and intelligent, data-driven decision-making in PCB recycling without the need to build expensive setups. 
PCB-Vision is a multi-scene RGB-HSI dataset, centered on PCBs, it unlocks a multitude of prospects for advancing computer vision techniques. Its potential extends beyond segmentation and object detection to applications like pansharpening and superpixel restoration in remote sensing. 
Our endeavor with this dataset is fortified by statistical analyses and crafted supervised learning ground truths for both RGB and hyperspectral data modalities. Through comprehensive numerical evaluations of segmentation models on these data types, we lay the foundation for innovative research endeavors and pioneering solutions to address the urgent challenges posed by E-waste recycling and support pathways to future concepts of more robot-aided and object-oriented E-waste component separation.
PCB-Vision can help identify the most valuable components of PCBs as we later present in the classes of interest, which can maximize the recovery of valuable materials and improve the efficiency of recycling processes \cite{eea2018}. 

PCB-Vision provides a platform for evaluating and comparing different PCB analysis methods. This benchmarking fosters innovation and collaboration in E-waste recycling by encouraging collaboration and innovation among researchers and industry. PCB-Vision pushes the E-waste recycling methods a step forward toward more efficient and environmentally friendly technologies by accelerating the development of non-invasive real-time analytical systems that meet the industrial demands for in-line data processing pipelines that assist subsequent tasks. This supports the Green Deal and the circular economy.

The subsequent sections of this paper are structured as follows: Section two provides a comprehensive overview of the dataset, addressing acquisition requirements, outlining the provided annotations, and presenting their associated statistics. Section three delves into the methodologies and models employed on the dataset, moreover, we detail the experimental procedures involved in training the models, along with the subsequent numerical evaluations. Section four discusses the forthcoming challenges and outlines the trajectory for future research directions. Finally, section five encapsulates our conclusions and summarizes the work's contributions.

\section{\textbf{Dataset Description}}
This section provides a comprehensive overview of the dataset, detailing the sensors employed for acquisition and presenting statistics related to the annotations. Figure \ref{fig:workflow} gives a glance at the PCB-Vision workflow. The conveyor belt industrial setup of the data acquisition shows the RGB and HSI cameras mounted on top of the belt scanning the PCBs and giving high spatial resolution images along with high spectral resolution HSI. The data is further preprocessed and normalized to fit the processing pipeline of DL models which after the training phase perform predictions to identify  valuable PCB components. 

\subsection{\textbf{Data Acquisition}}
\begin{figure}[h!]
  \centering
  \includegraphics[width=0.99\linewidth]{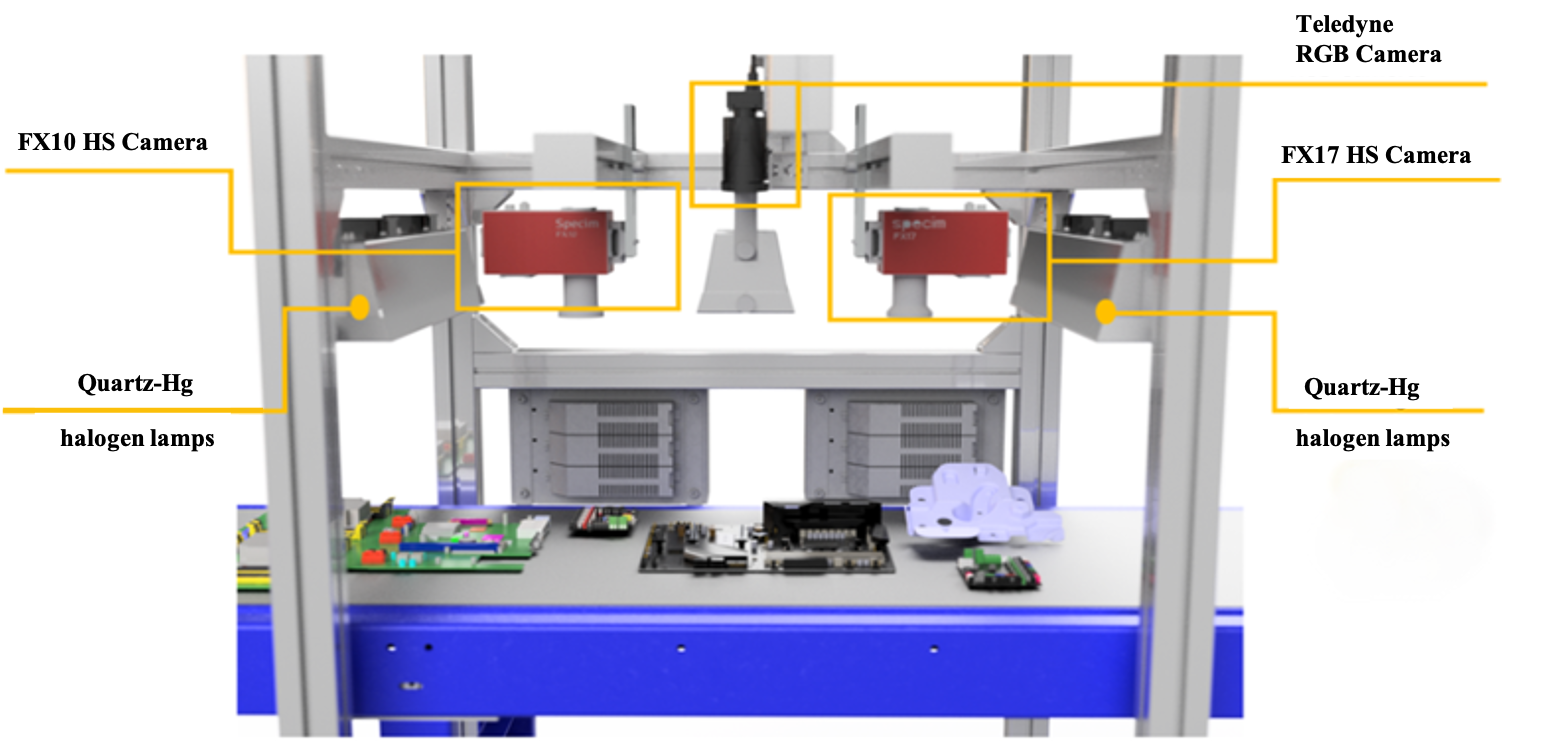}
  \caption{Acquisition setup at Helios Lab \cite{sudharshan2020object}}
  \label{fig:setup}
\end{figure}

The dataset encompasses RGB images captured by a Teledyne Dalsa Genie Nano-C4020 camera and HS data cubes obtained using a Specim FX10 spectrometer. The two sensors are positioned atop the conveyor belt in our Helios laboratory at the Helmholtz Institute Freiberg for Resource Technology (HIF) as shown in figure \ref{fig:setup}. The C4020 camera is characterized with a full-frame capture capability, while the FX10 spectrometer operates as a line scan device with high spatial resolution across the visible and near-infrared (VNIR) range. Illumination of the PCB stream was achieved using eight broad-band quartz-tungsten halogen lights. Table  \ref{tab:sensors-param} provides detailed acquisition parameters for the two cameras.
\begin{table}[!] 
  \caption{Acquisition cameras parameters}
  \centering
  \begin{tabular}{ccc}
    \toprule
    \textbf{Parameters} & \textbf{Teledyne Dalsa C4020 } & \textbf{Specim FX10} \\
    \midrule
    Spectral Range & RGB & 400-1000\,nm \\
    
    Spatial Resolution & 4000 $\cdot$ 3000\,px/\,frame & 1024 px/\,line \\
    
    Spectral Resolution & - & 5.5\,nm \\
    
    Bands & 3 & 224 \\
    
    Exposure Time& 1.5\,ms & 7.47\,ms \\
    
    Frame Rate & 8 Hz & 132\,fps \\
    
    Conveyor Speed & 10 cm/s & 10 cm/s \\
    
    \bottomrule
  \end{tabular}

  \label{tab:sensors-param}
\end{table}
Hyperspectral data acquisition with the FX10 was performed using the Specim Lumo Recorder software (Spectral Imaging Ltd., Oulu, Finland). Before capturing the PCBs cubes, a dark reference frame was acquired with the shutter closed. Subsequently, a white reference panel with 99\% reflectance was captured. Using the Hylite toolbox \cite{thiele2021multi}, we converted the raw hyperspectral datacubes to reflectance by applying the reference levels from the white and dark reference measurements, such that each pixel within the resulting cubes constitutes a vector representing the corrected reflectance spectra.

\begin{figure*}[h!]
    \begin{subfigure}[b]{0.45\textwidth}
        \centering
        \includegraphics[width=\textwidth]{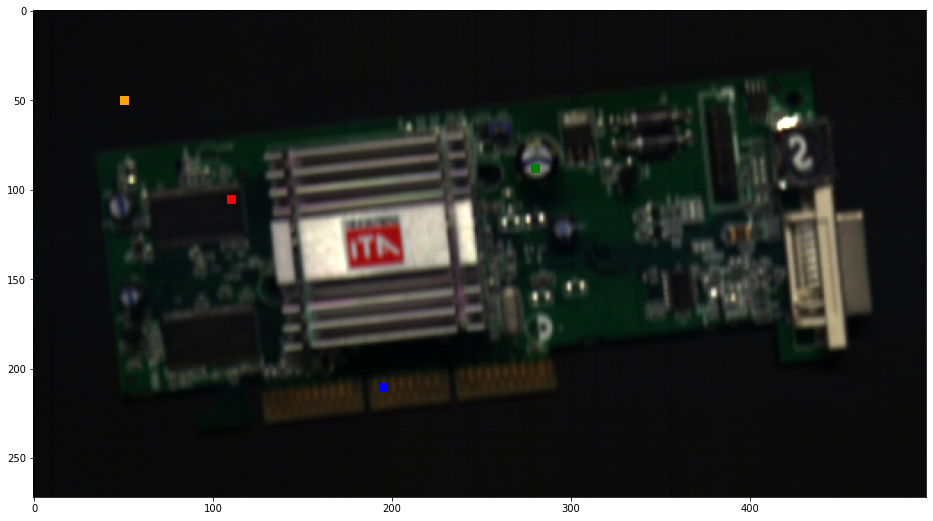}
        \caption{PCB1 True color representation HSI with the points of the spectral readings }
        \label{fig:pcb1-points}
    \end{subfigure}
    \hfill
    \begin{subfigure}[b]{0.45\textwidth}
        \centering
        \includegraphics[width=\textwidth]{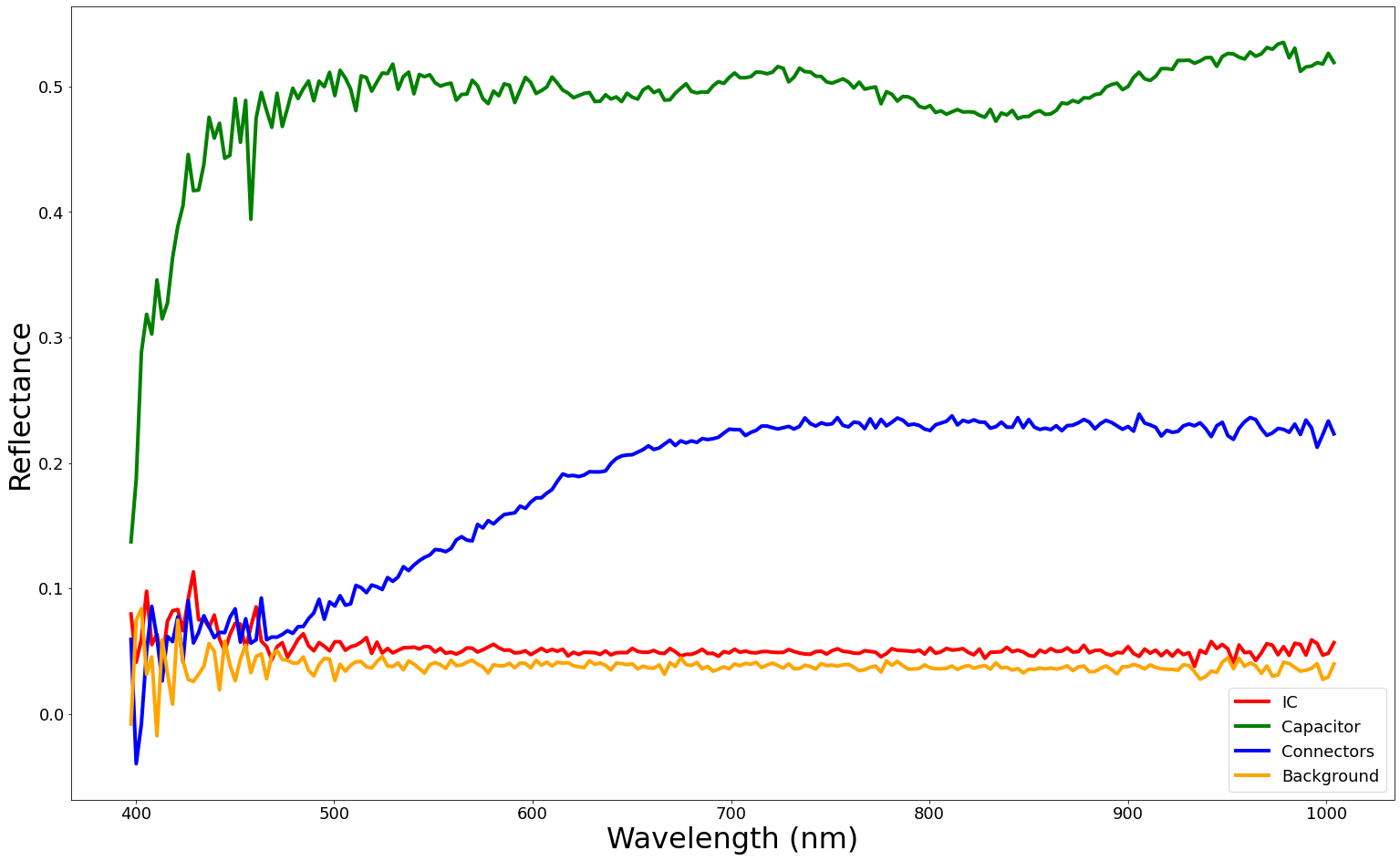}
        \caption{The four main classes spectra}
        \label{fig:spectra}
    \end{subfigure}
    \caption{Four spectra from PCB 1 HSI were captured from randomly selected points on the surface of our classes of interest, along with the spectra from the conveyor belt background. 'Conveyor belt' (orange), 'IC' (red), 'Capacitor' (green), and 'Connectors' (blue). }
    \label{fig:PCB1-spectra}
\end{figure*}

Figure \ref{fig:PCB1-spectra} presents PCB 1 true color representation of the HSI along the spectra of the classes of interest ('IC', 'Capacitor', 'Connectors'), in addition to the conveyor belt surface. The spectra were taken from a random point on the surface of the four objects.
From \ref{fig:spectra} we notice the spectra of class 'IC' are similar to the spectra of the conveyor belt. This is expected since the IC surface material and the conveyor belt material are similar black polymers, therefore they will reflect similar spectra in this range. We wanted to highlight this situation since this will confuse models that utilize spectral information only to do pixel-wise classification (segmentation) of those two classes. Further analysis of this problem is introduced in the experiments section.

For the annotation of hyperspectral cubes and the creation of pixel-wise classification ground truths, the Envi Classic (ENVI™ 5.1, Exelis Visual Information Solutions, Boulder, Colorado) software was employed. Simultaneously, RGB images were annotated using the Anylabeling \cite{vietanhdevanylabeling2023} annotation tool. These steps facilitated the generation of accurate and informative segmentation ground truths of three main components commonly found in every PCB, for subsequent computer vision tasks.

\subsection{\textbf{Annotations}}

Our research focuses on three fundamental objects commonly encountered in PCBs: Integrated Circuits (ICs), electrolytic capacitors, and connectors. The selection of ICs and connectors is driven by their significant economic value, while the inclusion of electrolytic capacitors is motivated by their potentially hazardous components. The accurate localization and detection of these components is a main task of our project's technical objectives. However, the tasks of our project 'Ramses' involve unique cases that demand a specialized type of segmentation ground truth different than the general segmentation ground truth that covers all the object's surface. For this purpose, an annotation ground truth named 'Monoseg' was created. Nonetheless, To enhance the dataset's versatility and contribution to the broader field of computer vision research, we also provide classic segmentation ground truths called 'General', making the dataset suitable for general-purpose tasks beyond the Ramses project. The main difference between the two segmentation types lies in the included object's surface area which is explained thoroughly further.
While it's conceivable to expand the number of object classes within the dataset, this process is subject to scalability considerations, ensuring a balanced and comprehensive representation of various PCB elements.

\subsubsection{\textbf{General Segmentation Annotations}}
\begin{figure}[h!]
    \begin{subfigure}[b]{0.45\textwidth}
        \centering
        \includegraphics[width=\textwidth]{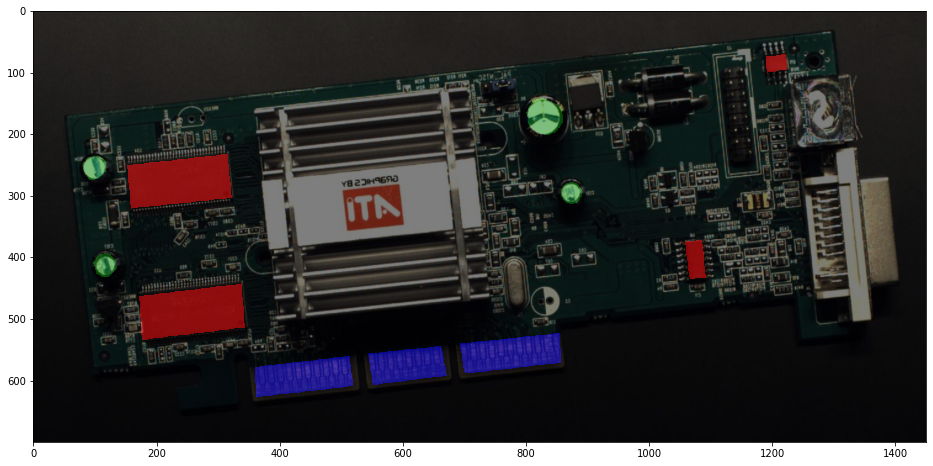}
        \caption{'General' ground truth projected on the RGB image}
        \label{fig:rgb-general}
    \end{subfigure}
    \hfill
    \begin{subfigure}[b]{0.45\textwidth}
        \centering
        \includegraphics[width=\textwidth]{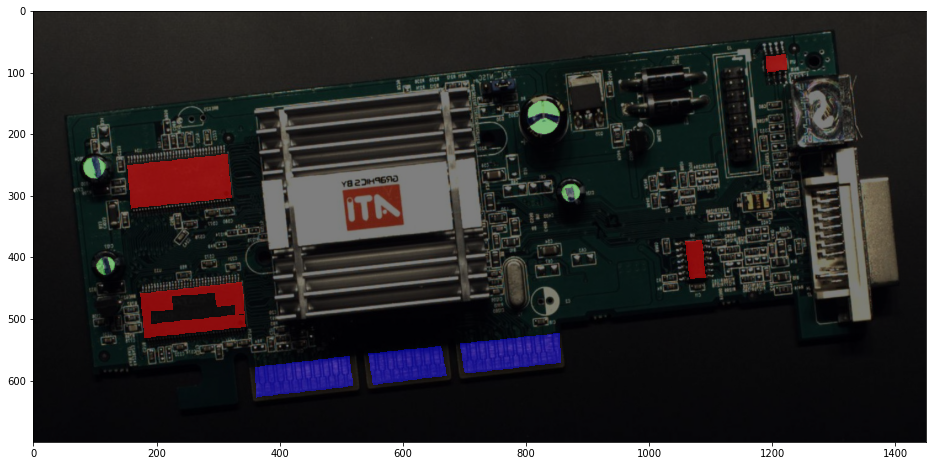}
        \caption{'Monoseg' ground truth projected on the RGB image}
        \label{fig:rgb-Monoseg}
    \end{subfigure}
    \caption{'General' and 'Monoseg' segmentation ground truth projected on PCB1 RGB image. 'IC' (red), 'Capacitor' (green), 'Connectors' (blue).}
    \label{fig:Anno-General}
\end{figure}

In this study, we define the 'General' segmentation ground truth as ground truth annotations encompassing all surfaces of the PCB components, regardless of the number of materials present on each surface. The resulting segmentation ground truth map for the PCB1 RGB image is depicted in  subfigure \ref{fig:rgb-general}. It can be seen how all the surface of each class of interest component is taken as ground truth. This is the main difference between the two annotation styles 'General' and 'Monoseg'.

\subsubsection{\textbf{Monoseg Annotations}}

In the context of the Ramses project, the segmentation ground truths adhere to a specific approach where only the primary material of an object's surface is labeled. This means excluding any additional markings, labels, or inscriptions present on PCB components' surfaces like labels that are printed on ICs surfaces. The rationale behind the 'Monoseg' ground truth style aligns with the use of point measurement sensors, such as XRF, and Raman spectroscopy that provides valuable chemical composition information about the reading point. In this approach, a point measurement is conducted within a given segment, and this measure must be taken where meaningful results are anticipated not where other labeling materials exist. Such obstructions of reading the wrong material point can lead to incorrect readings, potentially compromising the accuracy of the composition analysis. To mitigate this potential source of inaccuracies, the 'Monoseg' segmentation ground truth deliberately excludes all non-object-based materials.

\begin{figure}[h!]
    \begin{subfigure}[b]{0.45\textwidth}
        \centering
        \includegraphics[width=\textwidth]{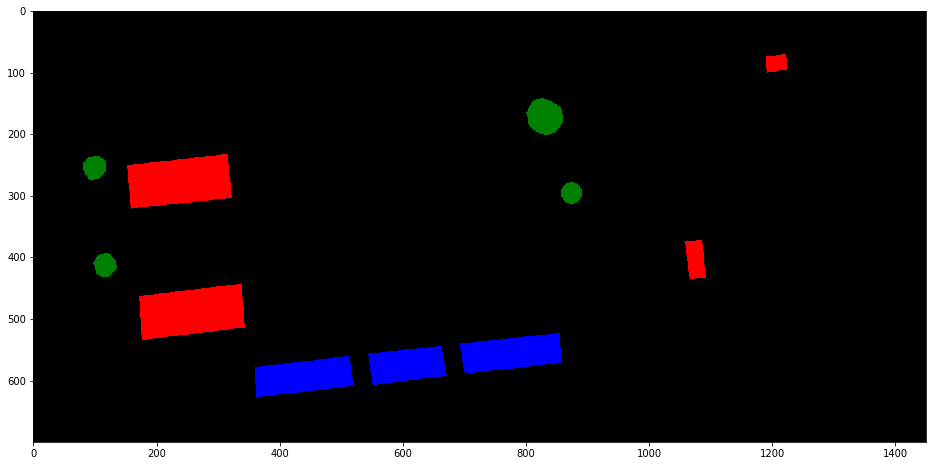}
        \caption{PCB 1 'General' ground truth}
        \label{fig:GeneralGT}
    \end{subfigure}
    \hfill
    \begin{subfigure}[b]{0.45\textwidth}
        \centering
        \includegraphics[width=\textwidth]{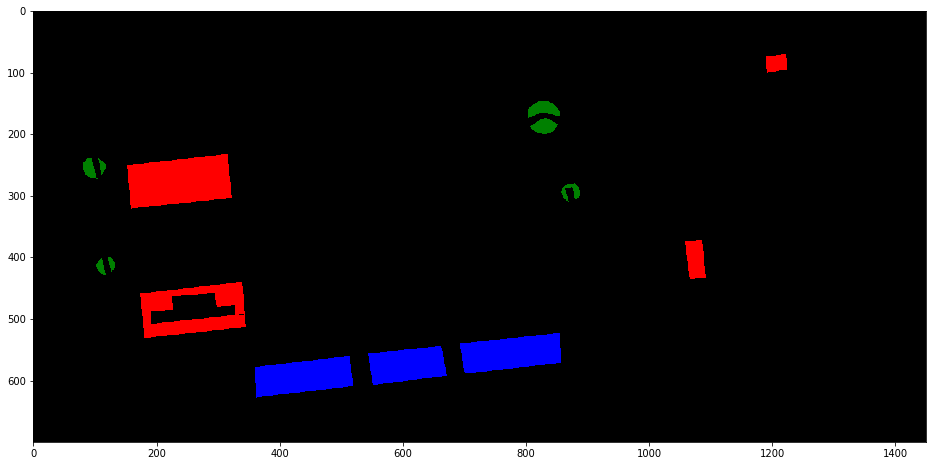}
        \caption{PCB 1 'Monoseg' ground truth}
        \label{fig:MonosegGT}
    \end{subfigure}
    \caption{'General' VS 'Monoseg' segmentation ground truth of PCB 1 RGB image. 'IC' (red), 'Capacitor' (green), 'Connectors' (blue).}
    \label{fig:GT_comparison}
\end{figure}

Subfigure \ref{fig:rgb-Monoseg} shows this annotation strategy on the PCB1 RGB image where 'Monoseg' ground truth considers only a specific part of the component surface, e.g., the surface of the bottom left IC is partially selected since the printing label on the surface is ignored.
For a better comparison of the two annotation styles, figure \ref{fig:GT_comparison} demonstrates the ground truth maps of the PCB 1 RGB image for each annotation style. 

The difference between the two annotation styles can be noticed in the bottom left IC and all the capacitors (green) where only the main material of the classes of interest are considered as targets.

\subsection{\textbf{Statistics}}
This section provides relevant statistics about the dataset and the ground truth containing the three classes of interest 'IC', 'Capacitor', and 'Connectors'. The two ground truth sets 'Monoseg' and 'General' have different sizes, the 'General' ground truth contains more segmented pixels for the 'IC' and 'Capacitor' classes. In this section, we present the statistics of the 'General' ground truth. However, although we provide two independent ground truths for both RGB and HSI, we highlight the ground truth of the hyperspectral ones, given its critical nature which plays an important role in selecting the training data for the HSI processing pipeline.

The dataset comprises 53 PCBs scanned using both hyperspectral and RGB cameras mounted on top of a black conveyor belt. The nature of this setup introduces the challenge of unwanted background (the black conveyor belt) surrounding the PCBs, significantly impacting various HSI preprocessing steps, such as data normalization, and dimensionality reduction, and processing steps like training State-Of-The-Art (SOTA) segmentation models. Moreover, the spectra of the conveyor belt closely resemble the spectra of the IC's dark plastic coating. To address this issue, we create PCB masks using the method proposed in \cite{arbash2023masking}, to help effectively eliminate the undesired background vectors and preserve only the PCB, our object of interest.

\begin{figure}[h!]
  \centering
  \includegraphics[width=0.5\linewidth]{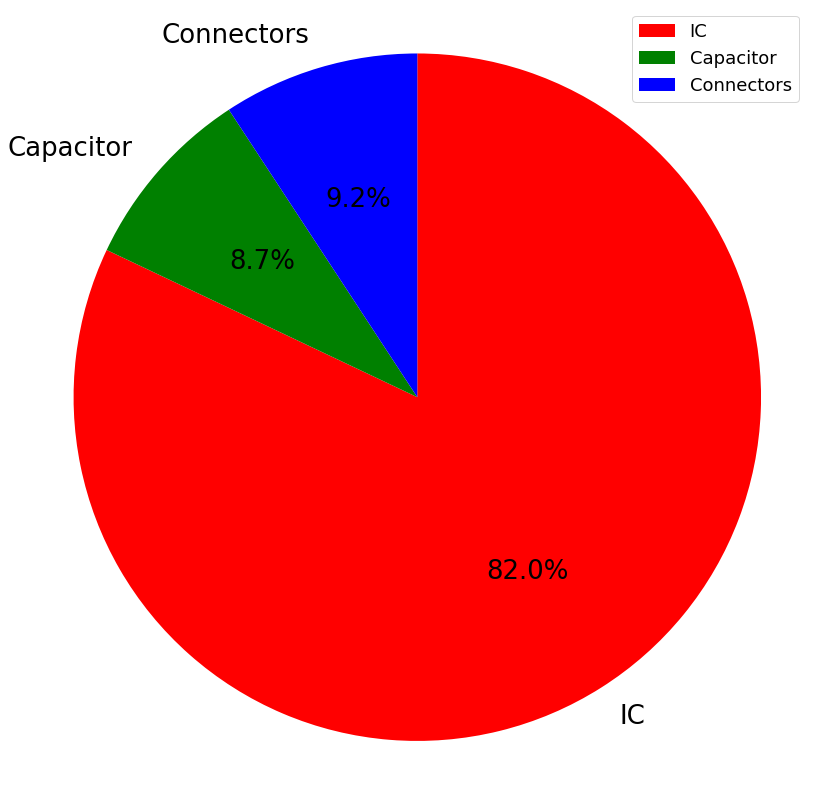}   
  \caption{Dataset classes percentages}
  \label{fig:class_dist}
\end{figure}

As mentioned previously, our dataset of 53 PCBs encompasses three classes: 'IC', 'Capacitor', and 'Connectors'. However, these classes are unbalanced, object-wise and pixel-wise across the dataset. Figure \ref{fig:class_dist} presents a pie chart illustrating the percentage of the mentioned classes pixels.
Figure \ref{fig:class_dist} reveals a highly imbalanced class case, where the 'IC' class (red) dominates the dataset, constituting 82.0\% of the ground truth, accounting for more than three-quarters of the entire dataset. Following this, the 'Connectors' class (blue) comprises 9.2\% of the dataset, while the 'Capacitor' class (green) represents 8.7\% of the dataset.

\begin{figure}[h!]
  \centering
  \includegraphics[width=0.75\linewidth]{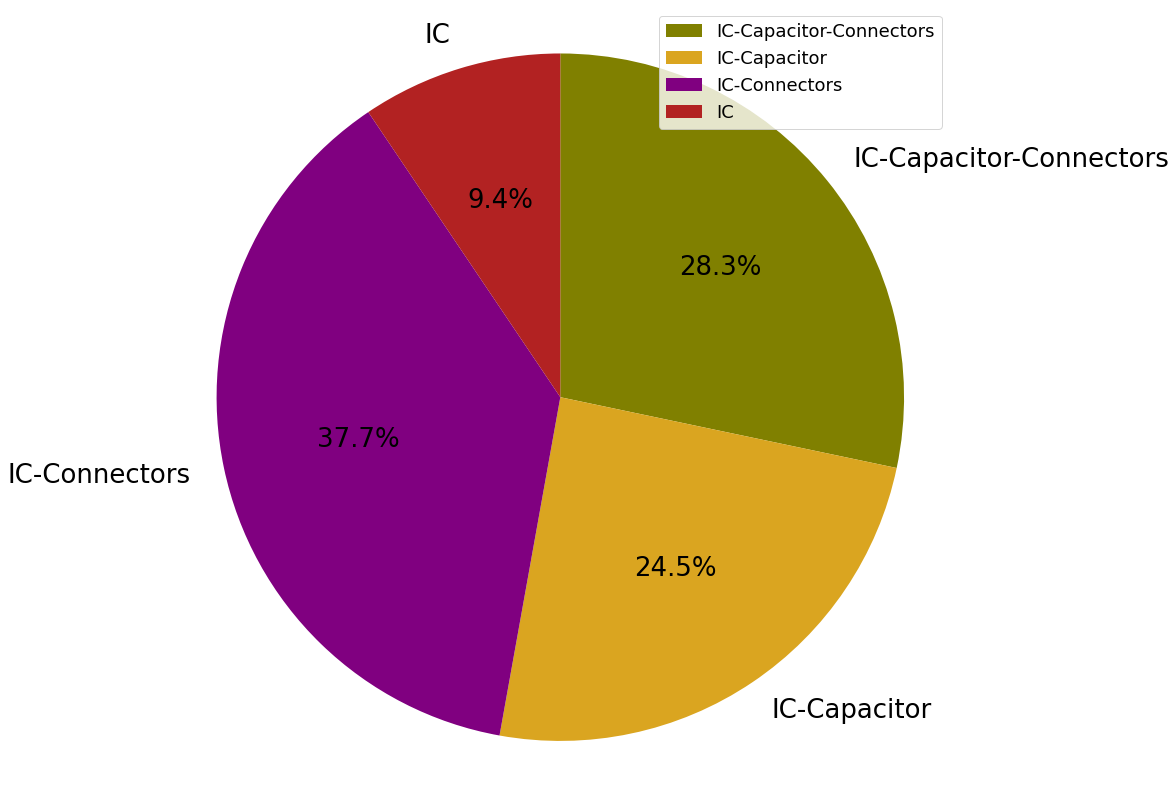}
  \caption{PCBs type based on the contained classes}
  \label{fig:pcbs_type}
\end{figure}

To gain a comprehensive understanding of the dataset and facilitate the subsequent HS cubes train-test split process, we categorized the PCBs based on the classes they contain. Figure \ref{fig:pcbs_type} presents a pie chart illustrating the percentage of PCB types based on their class composition.

As shown in figure \ref{fig:pcbs_type}, the most common PCB type consists of two classes only, namely "IC-Connectors" (purple), accounting for 37.7\% of the dataset. Following this, PCBs with all three classes, "IC-Capacitor-Connectors" (olive), constitute 28.3\% of the dataset size. PCBs containing another two classes, "IC-Capacitor" (dark yellow), represent 24.5\% of the dataset size. Finally, PCBs containing 'IC' class only (red) form 9.4\% of the dataset size.

\begin{figure}[h!]
    \begin{subfigure}[b]{0.45\textwidth}
        \centering
        \includegraphics[width=\textwidth]{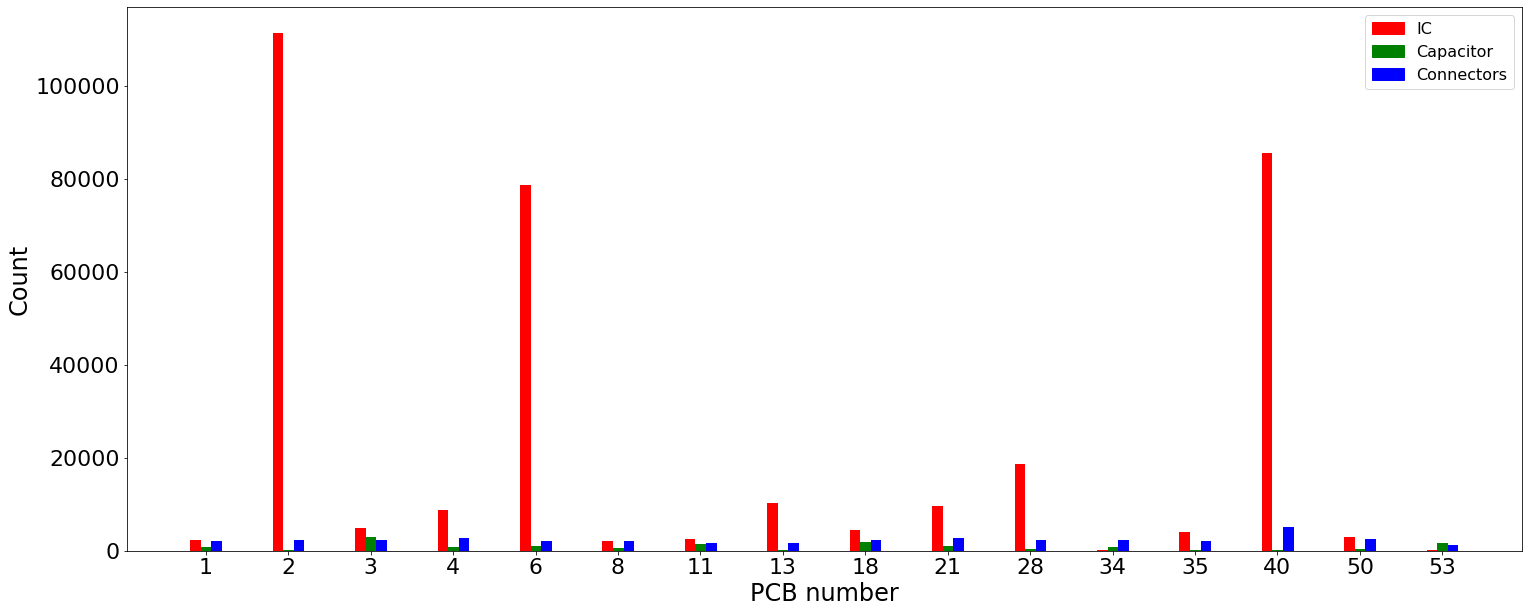}
        \caption{Histogram of (IC-Capacitor-Connectors) PCBs category}
        \label{fig:3classes}
    \end{subfigure}
    \hfill
    \begin{subfigure}[b]{0.45\textwidth}
        \centering
        \includegraphics[width=\textwidth]{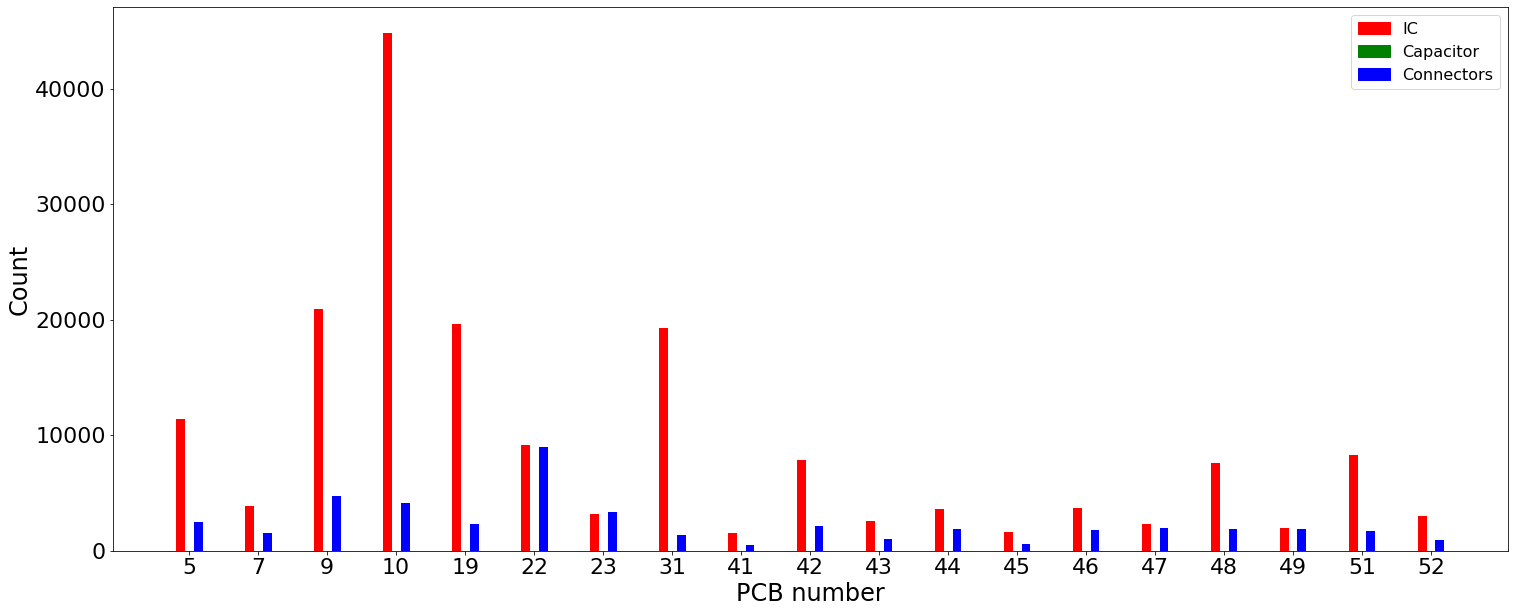}
        \caption{Histogram of 2 classes (IC-Connectors) PCB category}
        \label{fig:2classes_IC_con}
    \end{subfigure}
    \hfill
    \begin{subfigure}[b]{0.45\textwidth}
        \centering
        \includegraphics[width=\textwidth]{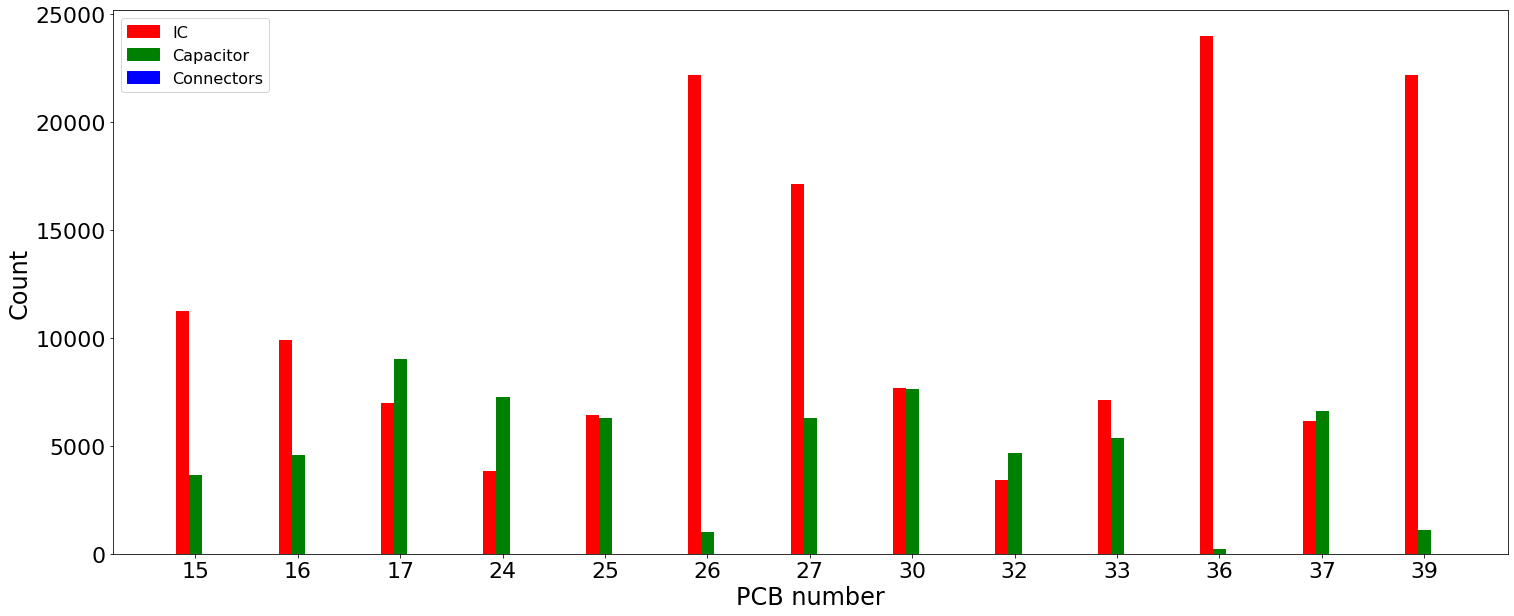}
        \caption{Histogram of 2 classes (IC-Capacitor) PCB category}
        \label{fig:2classes_IC_cap}
    \end{subfigure}
    \hfill
    \begin{subfigure}[b]{0.45\textwidth}
        \centering
        \includegraphics[width=\textwidth]{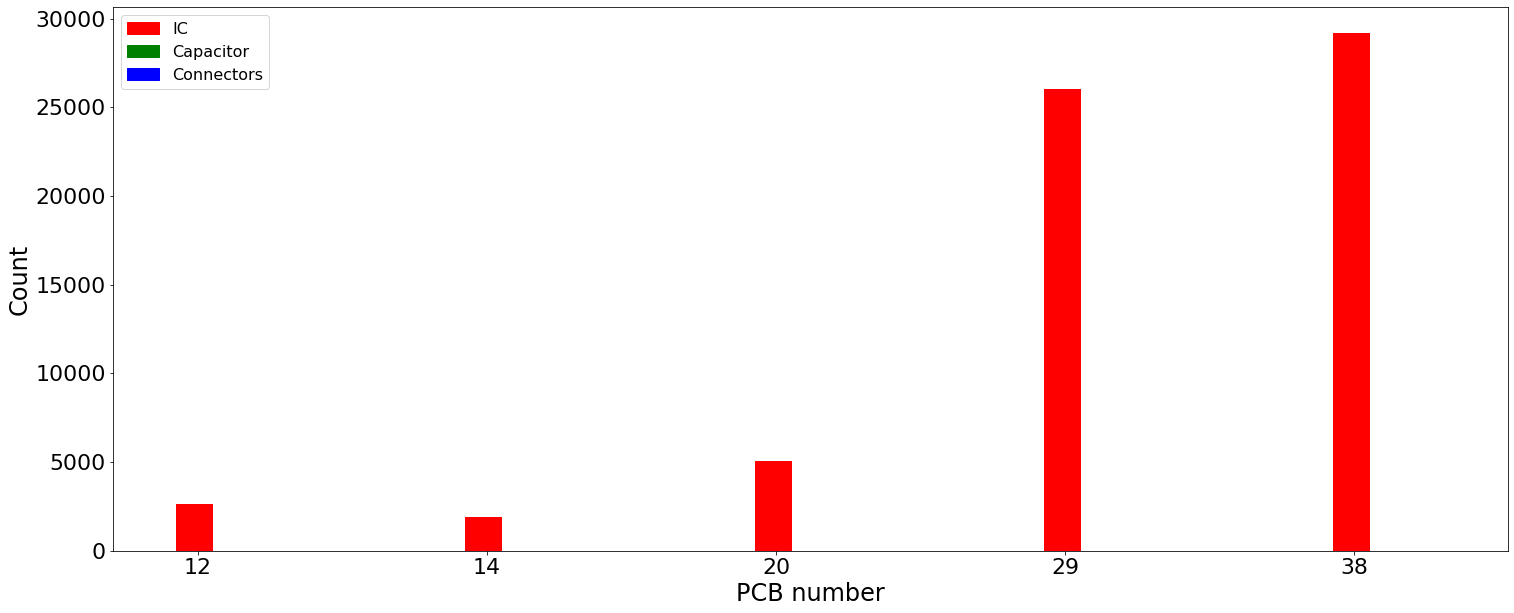}
        \caption{Histogram of 1 class (IC) PCB category}
        \label{fig:1class}
    \end{subfigure}
    \caption{Class distribution for each PCB in each PCB class category.}
    \label{fig:Histograms}
\end{figure}

To comprehensively analyze the class imbalance in our dataset corresponding to each PCB category as presented in figure \ref{fig:pcbs_type}, we provide histograms in figure \ref{fig:Histograms}. Each histogram illustrates the quantitative distribution of class pixels within every PCB of each category.

The four histograms presented in figure \ref{fig:Histograms} offer quantitative insights for the pixels' number (y-axis) in the HSI ground truth  for each PCB in the dataset (x-axis). The first histogram, denoted as figure \ref{fig:3classes}, corresponds to PCBs containing all three classes ('IC', 'Capacitor', and 'Connectors'), wherein the 'IC' class exhibits dominance, consistently observed in the other histograms as well. The second histogram, figure \ref{fig:2classes_IC_con}, pertains to PCBs containing both the 'IC' class and the 'Connectors' class. Similarly, the third histogram, figure \ref{fig:2classes_IC_cap}, represents PCBs with the 'IC' class and the 'Capacitor' class. Lastly, the fourth histogram, figure \ref{fig:1class}, is dedicated to PCBs containing only the 'IC' class. 
Those histograms provide essential insights for forming the right training set for our SOTA DL segmentation models on the HSI data since HSI model training is more sensitive than RGB due to the abundant information contained in the HSI that amplifies the class imbalance. It is common knowledge that DL models that are trained on biased datasets will give biased predictions, therefore the above-mentioned histograms helped build a hyperspectral training set as balanced as it could be. The hyperspectral PCB cubes forming the training set are 1, 3, 8, 11, 17, 22, 23, 24, 25, 32, 34, 44, 45, 47, 49, 50, 52, 53. The hyperspectral PCB cubes forming the validation set are 18, 37, 42. The rest of the cubes are used to create the test set.

\section{\textbf{Experiments}}
This section presents the segmentation experiments for the two ground truths 'General' and 'Monoseg', for both the HSI and the RGB data types using the raw version of the data without deploying any background (conveyor belt) masking technique to provide a benchmark point of PCB-Vision raw data. Five of the most famous segmentation models were used to benchmark the segmentation performance on the dataset. The following methodologies subsection elaborates more on those models.

\subsection{\textbf{Methodologies}}
In this study, we conducted comprehensive experiments to evaluate the performance of various semantic segmentation SOTA models including U-Net \cite{unet}, DeepLabv3+ \cite{deeplabv3+}, and Attention U-Net \cite{attention-unet}, among others. These models were chosen for their well-established capabilities in addressing semantic segmentation challenges in computer vision. The used models are: 

\subsubsection{Unet}
Unet is a widely used convolutional neural network (CNN) architecture for semantic segmentation tasks in computer vision. The network's unique design features a symmetric U-shaped encoder-decoder structure \cite{unet}. The Unet architecture efficiently captures multi-scale contextual information through skip connections, allowing precise segmentation of objects in images \cite{unet}. Its effectiveness and versatility have made Unet a popular choice for various segmentation applications.

\subsubsection{ResUnet}
ResUnet is a novel CNN architecture that combines the power of ResNet and Unet for semantic segmentation tasks in computer vision. The ResUnet model leverages residual connections from ResNet to facilitate efficient feature propagation during the encoder-decoder process \cite{resunet}. This fusion enables ResUnet to capture both local and global contextual information, enhancing segmentation accuracy and robustness. 

\subsubsection{Attention Unet}
Attention Unet is an innovative CNN architecture designed for precise semantic segmentation tasks in computer vision. The model integrates attention mechanisms within the standard Unet framework. By selectively attending to informative regions, Attention Unet achieves improved segmentation accuracy, particularly in cases where the target objects exhibit diverse appearances or complex structures \cite{attention-unet}. The network's attention mechanisms enable it to focus on relevant image regions, enhancing its ability to accurately delineate objects of interest, such as organs in medical imaging or objects in natural scenes.

\subsubsection{DeepLabv3+}
DeepLabv3+ is an advanced CNN architecture tailored for accurate semantic segmentation tasks in computer vision. The model employs an encoder-decoder structure with atrous separable convolutions \cite{deeplabv3+}. This design enables DeepLabv3+ to effectively capture multi-scale contextual information while preserving fine spatial details.

\subsubsection{LinkNet}
LinkNet is a CNN architecture specifically designed for efficient semantic segmentation. It introduces a novel encoding path, featuring residual-like connections called "link" connections, to preserve spatial information effectively \cite{linknet}. The link connections enable seamless integration of high-level features from deeper layers with low-level features from shallower layers, facilitating accurate segmentation.

Through meticulous evaluation, we investigated the numerical results of each model's inference on the test set, enabling a comprehensive comparison of their performance in accurately segmenting our classes of interest. 

\subsection{\textbf{Results: RGB}}

This subsection outlines the preprocessing steps applied to the RGB data and provides a numerical evaluation of the aforementioned DL models:
\begin{enumerate}
    \item \textbf{Train, validation, and test split}: the original RGB dataset comprises 53 images of the 53 PCBs. To facilitate effective model training, validation, and testing, we performed a split, allocating 62\% of the data for training, 16\% for validation, and 20\% for testing. This results in 33 images used for training, 9 for validation, and 11 for testing. The split was conducted randomly to ensure representative subsets in each category. We did not use the HSI training set built based on the statistics since the training data in the HSI are much higher than in RGB due to data abundance inherited in hyperspectral imaging consisting of 224 bands. The usage of 62\% of PCB-Vision data for training can yield better results.

    \item \textbf{Data augmentation}: is a major technique in DL that enhances models' performance by artificially expanding the training dataset \cite{data_aug}. By using the existing samples to create new synthetic data variants, data augmentation addresses the challenges of limited data availability and improves the generalization ability of DL models to unseen data \cite{data_aug}. Albumentations \cite{albumentation}, a powerful image augmentation Python library, was employed to apply various transformations to both RGB images and their corresponding masks. These transformations fall into two categories:
\begin{itemize}
    \item Spatial-level transformations (applied to both image and mask simultaneously):
\begin{itemize}
        \item Vertical flip
        \item Horizontal flip
        \item 40 degrees clockwise rotation
        \item 40 degrees anti-clockwise rotation
        \item RGB channel color shifting with a 25-shift limit for the red, green, and blue channels
        \item Transpose
        \item Shift scale rotate
\end{itemize}
    \item Pixel-level transformations (applied to the image without mask changes):
    \begin{itemize}
        \item Color jittering
        \item Channel shuffling
    \end{itemize}
\begin{itemize}
        \item Random snow
        \item One of: random brightness or random gamma
        \item One of: blur or motion blur
        \item One of: random contrast or hue saturation value
\end{itemize}

\end{itemize}

By introducing augmented data during training, the DL model increases its generalization ability, thereby improving performance across unseen data and diverse real-world scenarios. This approach mitigates overfitting risks, ensuring the model's efficacy in accurately segmenting objects under varying lighting conditions, viewpoints, and occlusion patterns \cite{data_aug2}. In the context of semantic segmentation, data augmentation is an indispensable strategy to boost model performance and ensure its practical applicability in various real-world scenarios \cite{data_aug}.

\item \textbf{Class imbalance mitigation}: While our dataset comprises three classes ('IC,' 'Capacitor,' and 'Connectors') another class named 'Others' is introduced for the model to classify everything that is not one of our classes of interest as 'Others'. This includes every other object that appears on the PCB surface, plus the undesired background (conveyor belt) as well. Class imbalance can impact the performance of machine learning models. To mitigate this, class weights were introduced to the loss function in order to highlight the error and updates for the classes of interest since they have fewer training samples than the 'Others' class.\\

\end{enumerate}

The following part contains two types of segmentation experiments, the first is done on the 'General' ground truth set, and the second is done with the 'Monoseg' type of ground truth. The five above-mentioned benchmark segmentation models are used for the semantic segmentation of the RGB PCBs.

While alternative hyperparameters were explored, the selected set presented in table \ref{tab:RGB-hyper} empirically demonstrated robust performance across all models, avoiding overfitting issues.

\begin{table}[h]
    \centering
    \caption{RGB images training hyperparameters}
    \label{tab:RGB-hyper}
    \begin{tabular}{cc}
         \toprule
         \textbf{Hyperparameter} & \textbf{Value} \\
         \midrule
         Image spatial resolution& 640x640x3 \\
         Batch size& 8 \\
         Optimizer & Adam \\
         Loss function& Weighted CCE \\
         Learning rate& 1e-4 \\
         Class weights& .1 - .8 - .85 - .9 \\
         Early stopping& 10 \\
         \bottomrule
    \end{tabular}
\end{table}

A noteworthy observation is that attempting a higher spatial image resolution (1280x1280x3) yielded improved results, despite practical constraints, particularly memory limitations, i.e., lack-of-memory errors that prevented universal application across all five models. Additionally, reducing the batch size below 8 negatively impacted performance, even with higher spatial resolutions.

\textbf{Note}: different hyperparameters were tested, resulting in varied outcomes for different models, mostly biased and overfitted performance. The chosen parameters strike a balance, showcasing good performance without encountering overfitting concerns for any of the models.

\paragraph{\textbf{'General' segmentation evaluation}}
Table \ref{tab:rgb-general} demonstrates the performance of the five models on the 'General' ground truths annotations.
\begin{table}
    \caption{Models' evaluation metrics on the RGB test set 'General' ground truth.}
        \begin{adjustbox}{width=0.48\textwidth}
    \centering
    \begin{tabular}{ccccccc} 
    \toprule
         \multicolumn{2}{c}{\textbf{Model}}&           \textbf{Metric $\backslash$ Class}&  \textbf{Others}&  \textbf{IC}   &  \textbf{Capacitor}   & \textbf{Connectors}\\ 
         \midrule
         \multicolumn{2}{c}{\textbf{Unet}}&  Precision&  0.96&  0.39&  0.74& 0.63\\ 
         \multicolumn{2}{c}{}&  Recall   &  0.99&  0.04&  0.69& 0.56\\ 
         \multicolumn{2}{c}{}&  F1 Score &  0.98&  0.07&  0.71& 0.60\\ 
         \multicolumn{2}{c}{}&  IOU      &  0.96&  0.03&  0.55& 0.42\\ 
         \multicolumn{2}{c}{\textbf{Attention Unet}}&  Precision&  0.98&  0.73&  0.83& 0.39\\ 
         &  &  Recall   &  0.99&  0.55&  0.73& 0.99\\ 
         &  &  F1 Score &  0.98&  0.63&  0.78& 0.56\\ 
         &  &  IOU      &  0.97&  0.46&  0.64& 0.39\\ 
 \multicolumn{2}{c}{\textbf{ResUnet}}& Precision& 0.97& 0.25& 0.48&0.11\\ 
 & & Recall   & 0.97& 0.11& 0.86&0.63\\ 
 & & F1 Score & 0.97& 0.16& 0.62&0.20\\ 
 & & IOU      & 0.94& 0.09& 0.45&0.11\\ 
 \multicolumn{2}{c}{\textbf{DeepLabv3+}}& Precision& 0.98& 0.58& 0.84&0.81\\ 
 & & Recall   & 0.98& 0.66& 0.70&0.68\\ 
 & & F1 Score & 0.98& 0.62& 0.76&0.74\\ 
 & & IOU      & 0.97& 0.45& 0.62&0.59\\ 
 \multicolumn{2}{c }{\textbf{LinkNet}}& Precision& 0.98& 0.39& 0.63&0.72\\ 
 & & Recall   & 0.97& 0.61& 0.80&0.92\\ 
 & & F1 Score & 0.98& 0.48& 0.70&0.81\\ 
 & & IOU      & 0.95& 0.31& 0.54&0.68\\ 
 \bottomrule
    \end{tabular}
    \end{adjustbox}
    \label{tab:rgb-general}
\end{table}
Based on table \ref{tab:rgb-general} it can be concluded that the DeepLabv3+ and Attention UNet models stand out as the effective models among those five models exhibiting strong performance across precision, recall, and F1 Score for all classes, indicating a robust overall segmentation capability.

\begin{figure*}[t]
  \centering
  \includegraphics[width=0.7\textwidth]{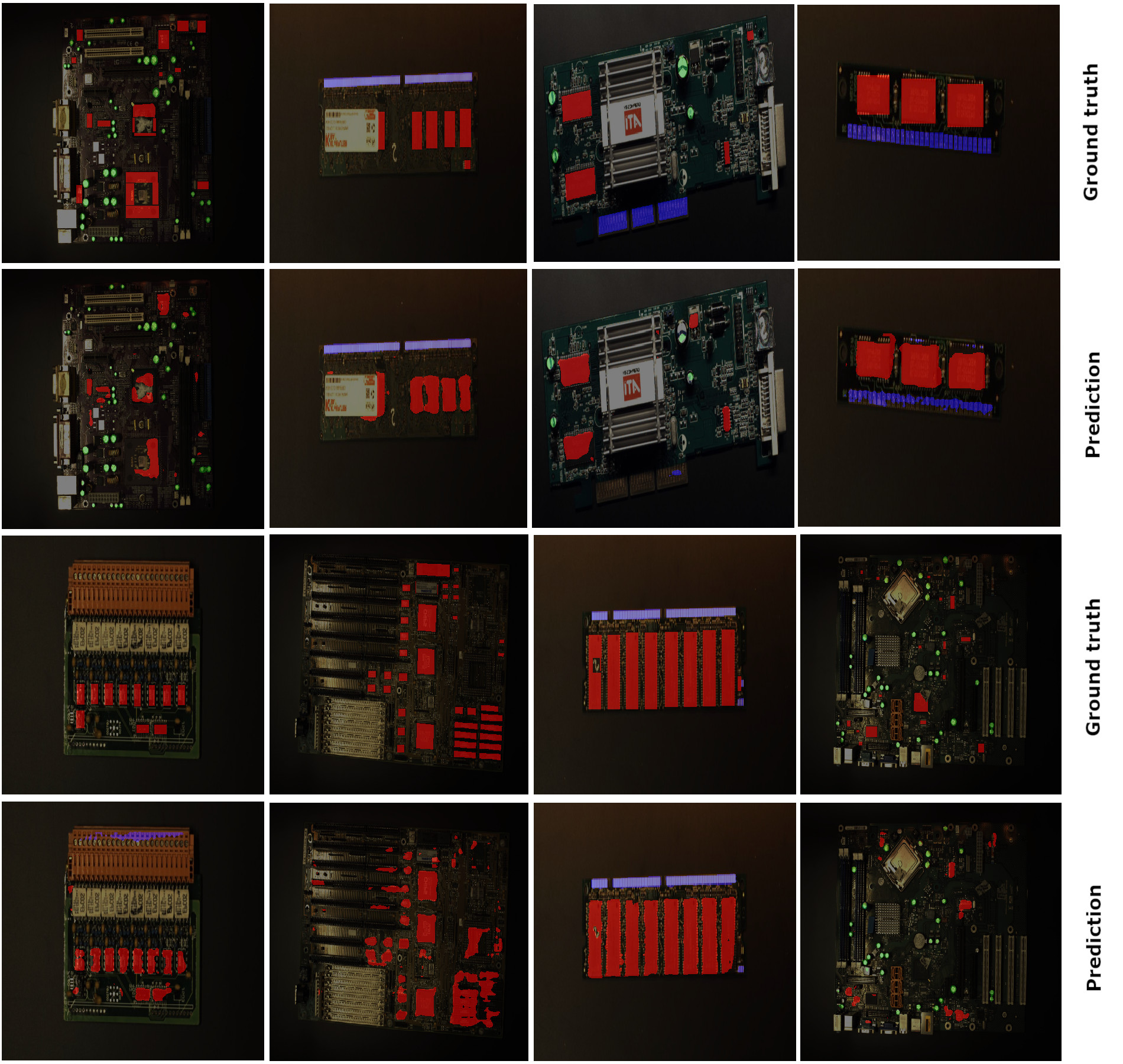}  
  \caption{Visual comparison of DeepLabv3+ predictions and ground truths on eight PCBs from the test set.}
  \label{fig:RGB Predictions}
\end{figure*}

Figure \ref{fig:RGB Predictions} presents the prediction performance of the DeepLabv3+ on some of the test images.

The quality of the segmentation prediction can be assessed by comparing the predicted masks to the ground truth masks. A good segmentation prediction should have a high overlap with the ground truth mask, and simultaneously be free of noise and artifacts.

From figure \ref{fig:RGB Predictions}  it can be seen that the DeepLabv3+ model performs well on the three classes, with a high overlap between the predicted and ground truth masks. However, there are some cases where the model fails. Here is more expansion on the overall performance of DeepLabv3+:
\begin{itemize}
    \item The model performs very well on the 'IC' class, with a high overlap between the predicted and ground truth masks.
    \item The model also performs well on the 'Capacitor' class, but there are some cases where the model misses some of the 'Capacitor' pixels, e.g., the last PCB (bottom right).
    \item The model performs worst on the 'Connectors' class, missing some of the Connector pixels, e.g., the third PCB from the left.
    \item The model tends to make more mistakes in images with complex backgrounds.
\end{itemize}

Overall, the DeepLabv3+ model is a promising model for PCB segmentation. However, it is important to note that no model is perfect, and there will always be some cases where the model makes mistakes, especially in imbalanced class scenarios our PCB-Vision contains.

\paragraph{\textbf{'Monoseg' segmentation evaluation}}
Table \ref{tab:rgb-Monoseg} demonstrates the performance of the five segmentation models on the 'Monoseg' ground truth annotations.
\begin{table}
    \caption{Models' evaluation metrics on the RGB test set 'Monoseg' ground truth.}
        \begin{adjustbox}{width=0.48\textwidth}
    \centering
    \begin{tabular}{ccccccc} 
    \toprule
         \multicolumn{2}{c}{\textbf{Model}}&           \textbf{Metric $\backslash$ Class}&  \textbf{Others}&  \textbf{IC}   &  \textbf{Capacitor}   & \textbf{Connectors}\\ 
         \midrule
         \multicolumn{2}{c}{\textbf{Unet}}&  Precision&  0.98&  0.66&  0.58& 0.57\\ 
         \multicolumn{2}{c}{}&  Recall   &  0.99&  0.38&  0.90& 0.95\\ 
         \multicolumn{2}{c}{}&  F1 Score &  0.99&  0.48&  0.71& 0.71\\ 
         \multicolumn{2}{c}{}&  IOU      &  0.98&  0.32&  0.55& 0.56\\ 
         \multicolumn{2}{c}{\textbf{Attention Unet}}&  Precision&  0.98&  0.57&  0.39& 0.25\\ 
         &  &  Recall   &  0.98&  0.35&  0.96& 0.94\\ 
         &  &  F1 Score &  0.98&  0.43&  0.56& 0.39\\ 
         &  &  IOU      &  0.97&  0.27&  0.39& 0.24\\ 
 \multicolumn{2}{c}{\textbf{ResUnet}}& Precision& 0.98& 0.26& 0.37&0.15\\ 
 & & Recall   & 0.98& 0.09& 0.84&0.82\\ 
 & & F1 Score & 0.97& 0.14& 0.53&0.25\\ 
 & & IOU      & 0.96& 0.07& 0.37&0.14\\ 
 \multicolumn{2}{c}{\textbf{DeepLabv3+}}& Precision& 0.99& 0.51& 0.49&0.58\\ 
 & & Recall   & 0.98& 0.66& 0.92&0.96\\ 
 & & F1 Score & 0.99& 0.58& 0.64&0.72\\ 
 & & IOU      & 0.97& 0.40& 0.47&0.57\\ 
 \multicolumn{2}{c}{\textbf{LinkNet}}& Precision& 0.99& 0.47& 0.49&0.86\\ 
 & & Recall   & 0.98& 0.61& 0.84&0.97\\ 
 & & F1 Score & 0.99& 0.53& 0.62&0.91\\ 
 & & IOU      & 0.97& 0.36& 0.44&0.83\\ 
 \bottomrule
    \end{tabular}
  \end{adjustbox}
    \label{tab:rgb-Monoseg}
\end{table}

Based on table \ref{tab:rgb-Monoseg}, DeepLabv3+ exhibits strong overall performance, excelling in both precision and recall. This suggests its effectiveness in capturing true positives while minimizing false positives and false negatives. Additionally, LinkNet demonstrates high precision across all classes, indicating accurate positive predictions. It achieves a moderate performance in terms of recall, precision, and F1 Score, making it another robust choice to consider for this particular type of ground truth.

\subsection{\textbf{Results: HSI}}
In the context of hyperspectral data, several preprocessing steps are undertaken to enhance the quality and relevance of the dataset. These steps are crucial for ensuring that the hyperspectral information is effectively leveraged for subsequent model training.

\begin{enumerate}
    \item \textbf{Data normalization}: The spectra in the hyperspectral data are normalized using information from the dark acquisition and white reference panel. This normalization process ensures that the spectral values fall within the standardized range of zero to one.
    \item \textbf{Data limiting}: Post-normalization, some values might still fall outside the expected range due to factors like noise or the presence of materials with higher reflectance than the white reference panel, e.g., shiny metals (heat sink) on the PCB surface. To address this, values exceeding the range are limited to 1.0, while negative values are set to 0.0.
    \item \textbf{Train, validation, test split}: Given the imbalanced nature of classes in hyperspectral data, a manual split is performed to ensure a balanced representation in the training set. This is particularly important to mitigate the impact of amplified class imbalance in HSI on model training. As stated in the statistics, the dataset is divided into training (56\%), validation (5\%), and test (39\%) sets. 
Figures \ref{fig:HSI training}, \ref{fig:HSI Validation}, \ref{fig:HSI Testing} demonstrates the split of the PCB-Vision HS data cubes into the three training, validation, and testing sets respectively.
\begin{figure*}[t]
  \centering
  \includegraphics[width=0.8\textwidth]{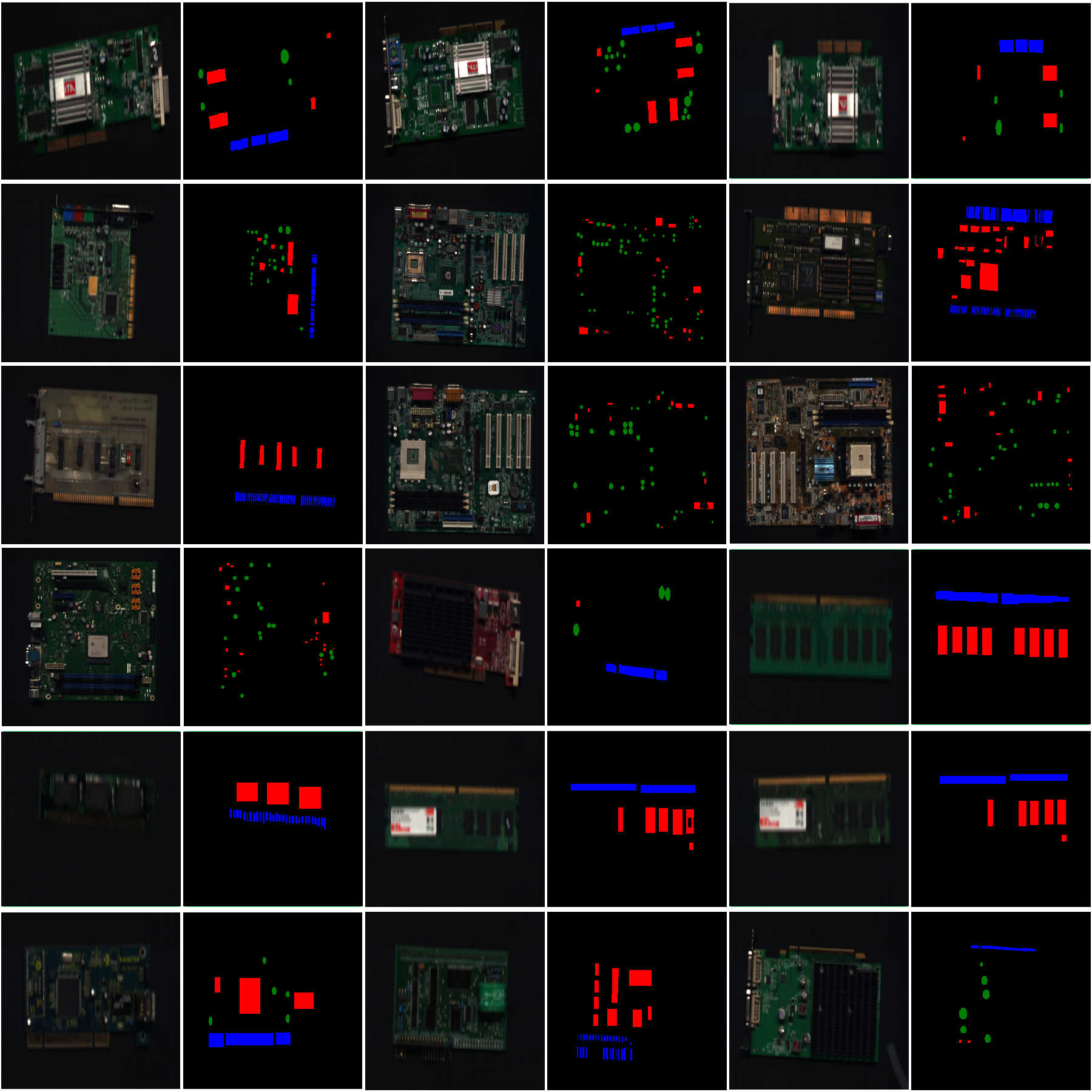}  \caption{HSI training set.}
  \label{fig:HSI training}
\end{figure*}
\begin{figure}
    \centering
    \includegraphics[width=0.8\linewidth]{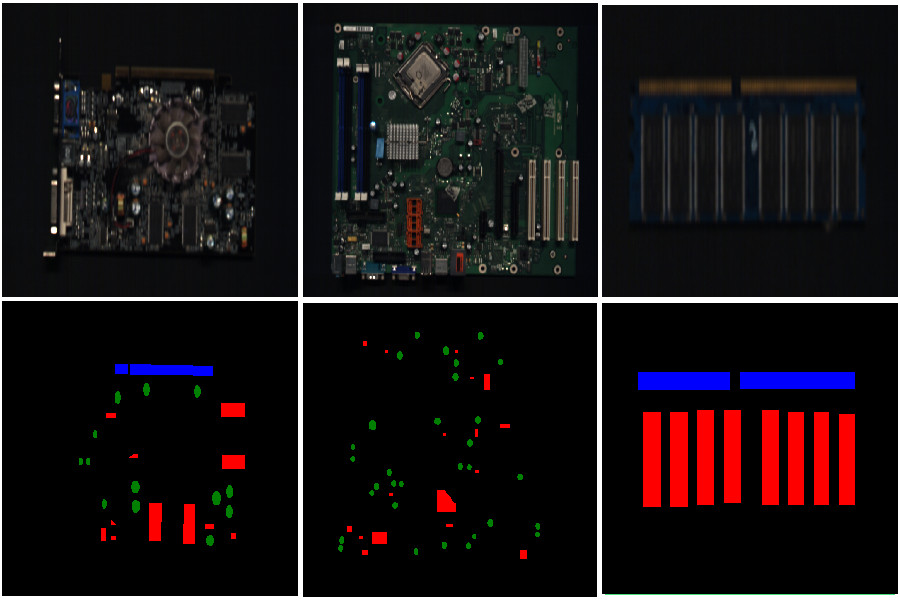}
    \caption{HSI validation set.}
    \label{fig:HSI Validation}
\end{figure}
\begin{figure}
    \centering
    \includegraphics[width=0.8\linewidth]{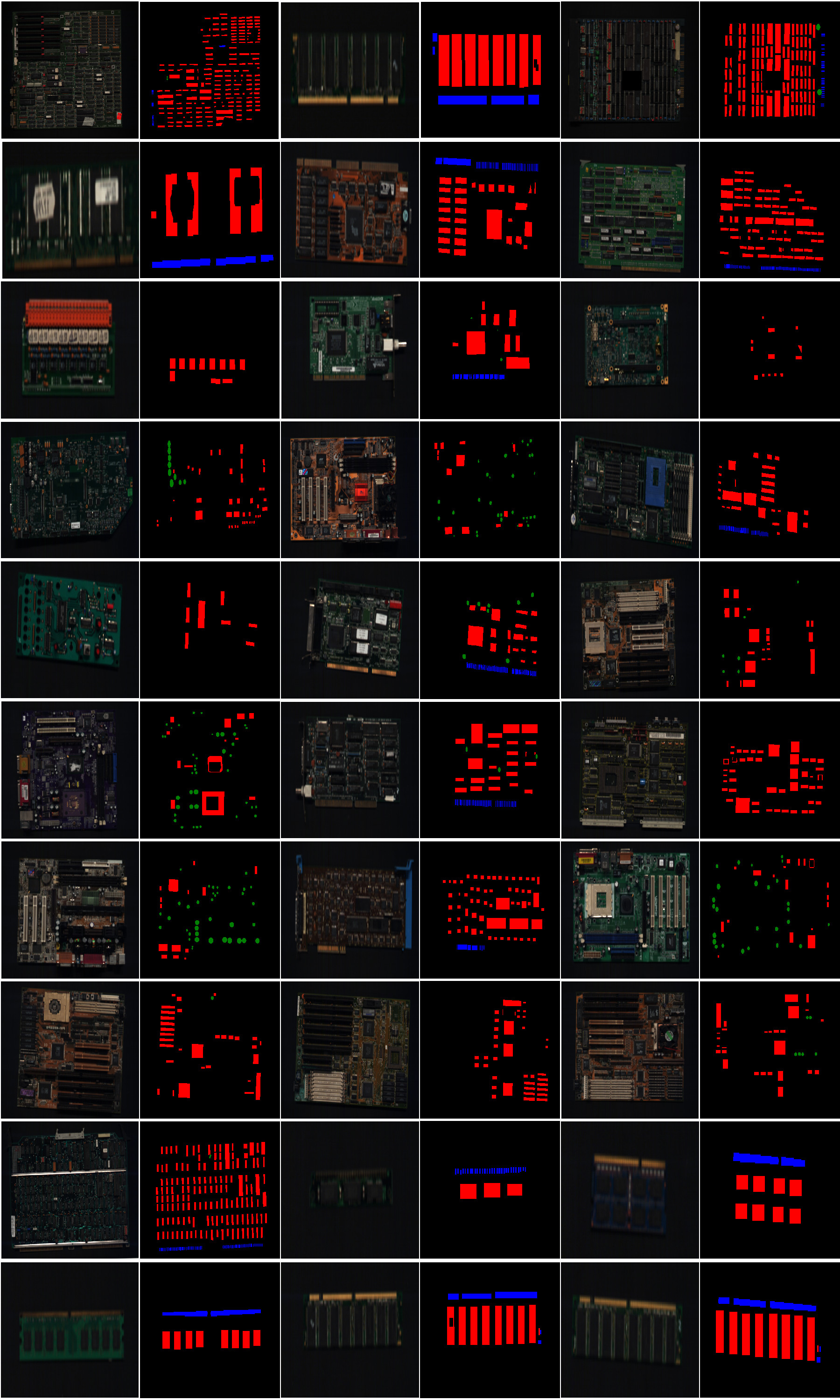}
    \caption{HSI testing set.}
    \label{fig:HSI Testing}
\end{figure}

    \item \textbf{Data augmentation}: Data augmentation techniques, including random rotation (clockwise and counterclockwise), translation (vertical and horizontal), as well as horizontal and vertical flipping, are applied. These augmentations contribute to the model's ability to generalize and perform well on unseen data.
    \item \textbf{Class imbalance mitigation}: To address class imbalance, class weights are incorporated into the loss function during model training. This ensures that the model gives due consideration to classes with fewer samples, thereby improving its ability to recognize and classify the classes of interest over the 'Others' class.
\end{enumerate}

Results are split into two main categories depending on the data type. We experimented with two types of data, the first is by reducing the dimensionality of the hyperspectral data cubes using Principal Component Analysis (PCA) utilizing the first three components. The second type of data was patches of the raw hyperspectral cubes. 
We ran two segmentation experiments on each of these two main types of data, one for the 'General' ground truths and one for the 'Monoseg' ground truths.

\subsubsection{\textbf{HSI - PCA}}
\begin{figure}
    \centering
    \includegraphics[width=.9\linewidth]{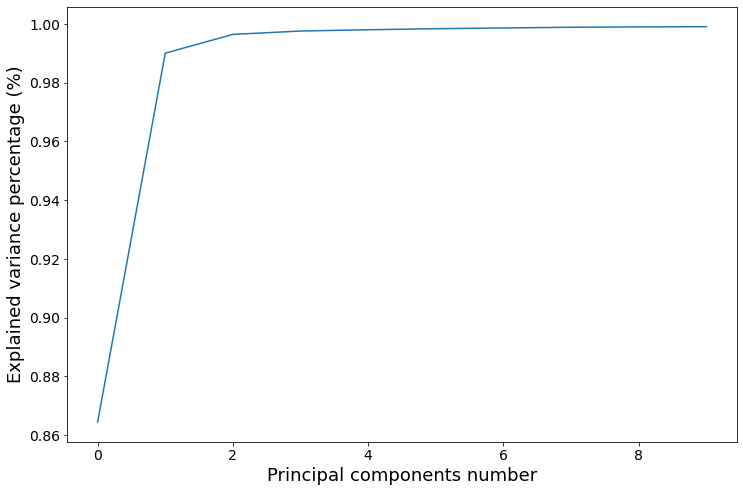}
    \caption{PCA explained variance}
    \label{fig:pca-var}
\end{figure}

Our motivation behind using PCA came from the two features our hyperspectral cubes have, the spectral and the spatial features, therefore we extracted the spectral features using PCA and the spatial features using the benchmark segmentation models in an attempt to combine the merits of the two methods.
The first three principal components were chosen since they cover up to 99\% of the variance in the data as can be seen in figure \ref{fig:pca-var}.

To augment the training set, we applied eight spatial-level augmentation techniques, enhancing the diversity of deep learning segmentation models input. These augmentations included clockwise and counterclockwise rotation, vertical and horizontal transition (positive and negative), and vertical and horizontal flipping. This resulted in a total of 126 PCA data samples for training. The hyperparameters that yielded the better results which were used for training the segmentation models on the PCA data are outlined in table \ref{tab:PCA-hyper}.

\begin{table}
    \centering
    \caption{PCA data training hyperparameters}
    \label{tab:PCA-hyper}
    \begin{tabular}{cc}
    \toprule
         Hyperparameter& Value\\
         \midrule
         Image resolution& 640x640x3\\
         Batch size& 8\\
         Optimizer& Adam\\
         Loss function& Weighted CCE\\
         Learning rate& 1e-4\\
 Class weights&.1 - .7 - .95 - .8\\
 Early stopping&20\\
 \bottomrule
    \end{tabular}
    
\end{table}
\paragraph{\textbf{PCA 'General' segmentation}}
Table \ref{tab:PCA-general} shows the numerical evaluation of the five benchmark models on the PCA data with the 'General' segmentation ground truth.
\begin{table}
    \caption{Models' evaluation metrics on the HSI PCA test set 'General' ground truth.}
    \begin{adjustbox}{width=0.48\textwidth}
    \centering
    \begin{tabular}{ccccccc} 
    \toprule
         \multicolumn{2}{c}{\textbf{Model}}&           \textbf{Metric $\backslash$ Class}&  \textbf{Others}&  \textbf{IC}   &  \textbf{Capacitor}   & \textbf{Connectors}\\ 
         \midrule 
         \multicolumn{2}{c}{\textbf{Unet}}&  Precision&  0.95&  0.65&  0.42& 0.11\\ 
         \multicolumn{2}{c}{}&  Recall   &  0.98&  0.33&  0.22& 0.10\\ 
         \multicolumn{2}{c}{}&  F1 Score &  0.96&  0.44&  0.29& 0.11\\ 
         \multicolumn{2}{c}{}&  IOU      &  0.93&  0.28&  0.17& 0.06\\ 
         \multicolumn{2}{c}{\textbf{Attention Unet}}&  Precision&  0.95&  0.51&  0.66& 0.11\\ 
         &  &  Recall   &  0.97&  0.34&  0.11& 0.20\\ 
         &  &  F1 Score &  0.96&  0.41&  0.19& 0.14\\ 
         &  &  IOU      &  0.92&  0.26&  0.11& 0.08\\ 
 \multicolumn{2}{c}{\textbf{ResUnet}}& Precision& 0.96& 0.27& 0.33&0.05\\ 
 & & Recall   & 0.87& 0.57& 0.24&0.14\\ 
 & & F1 Score & 0.91& 0.37& 0.28&0.07\\ 
 & & IOU      & 0.84& 0.22& 0.16&0.04\\ 
 \multicolumn{2}{c}{\textbf{DeepLabv3}+}& Precision& 0.94& 0.77& 0.54&0.47\\ 
 & & Recall   & 0.99& 0.25& 0.32&0.64\\ 
 & & F1 Score & 0.97& 0.38& 0.40&0.54\\ 
 & & IOU      & 0.93& 0.24& 0.25&0.37\\ 
 \multicolumn{2}{c}{\textbf{LinkNet}}& Precision& 0.95& 0.54& 0.13&0.40\\ 
 & & Recall   & 0.98& 0.30& 0.22&0.32\\ 
 & & F1 Score & 0.96& 0.39& 0.16&0.36\\ 
 & & IOU      & 0.92& 0.24& 0.09&0.22\\ 
  \bottomrule
    \end{tabular}
   \end{adjustbox}
    \label{tab:PCA-general}
\end{table}
From table \ref{tab:PCA-general} we conclude that as in the RGB cases, Deeplabv3+ exhibits strong performance across precision, recall, and F1 score for all classes, indicating a robust overall segmentation capability.

\paragraph{\textbf{PCA 'Monoseg' segmentation}}
Table \ref{tab:PCA-Monoseg} shows the numerical evaluation of the five benchmark models on the PCA data with the 'Monoseg' segmentation ground truth.
\begin{table}
\caption{Models' evaluation metrics on the HSI PCA test set 'Monoseg' ground truth.}
\begin{adjustbox}{width=0.48\textwidth}
    \centering
    \begin{tabular}{ccccccc} 
    \toprule
         \multicolumn{2}{c}{\textbf{Model}}&           \textbf{Metric} $\backslash$ \textbf{Class}&  \textbf{Others}&  \textbf{IC}   &  \textbf{Capacitor}   & \textbf{Connectors}\\ 
         \midrule 
         \multicolumn{2}{c}{\textbf{Unet}}&  Precision&  0.96&  0.52&  0.55& 0.06\\ 
         \multicolumn{2}{c}{}&  Recall   &  0.97&  0.27&  0.19& 0.20\\ 
         \multicolumn{2}{c}{}&  F1 Score &  0.97&  0.35&  0.29& 0.10\\ 
         \multicolumn{2}{c}{}&  IOU      &  0.94&  0.21&  0.17& 0.05\\ 
         \multicolumn{2}{c}{\textbf{Attention Unet}}&  Precision&  0.97&  0.37&  0.32& 0.06\\ 
         &  &  Recall   &  0.96&  0.44&  0.31& 0.09\\ 
         &  &  F1 Score &  0.96&  0.41&  0.31& 0.07\\ 
         &  &  IOU      &  0.93&  0.25&  0.19& 0.04\\ 
 \multicolumn{2}{c}{\textbf{ResUnet}}& Precision& 0.97& 0.14& 0.12&0.02\\ 
 & & Recall   & 0.82& 0.50& 0.34&0.15\\ 
 & & F1 Score & 0.88& 0.21& 0.18&0.04\\ 
 & & IOU      & 0.79& 0.12& 0.10&0.02\\ 
 \multicolumn{2}{c}{\textbf{DeepLabv3+}}& Precision& 0.96& 0.58& 0.41&0.35\\ 
 & & Recall   & 0.99& 0.16& 0.35&0.51\\ 
 & & F1 Score & 0.97& 0.25& 0.38&0.42\\ 
 & & IOU      & 0.95& 0.14& 0.23&0.26\\ 
 \multicolumn{2}{c}{\textbf{LinkNet}}& Precision& 0.97& 0.30& 0.07&0.30\\ 
 & & Recall   & 0.91& 0.52& 0.52&0.59\\ 
 & & F1 Score & 0.94& 0.38& 0.13&0.40\\ 
 & & IOU      & 0.98& 0.23& 0.07&0.25\\ 
 \bottomrule
    \end{tabular}
    \end{adjustbox}
    \label{tab:PCA-Monoseg}
\end{table}
Table \ref{tab:PCA-Monoseg} demonstrates that Attention U-Net and DeepLabv3+ exhibited comparable performances across multiple classes, except for the 'Connectors' class, where DeepLabv3+ achieved superior performance.

\subsubsection{\textbf{HSI - Patches} }
In this part, patches of raw data were fed to the DL models, in an attempt to construct an end-to-end model capable of discerning both spectral and spatial features promising to the segmentation of the four classes. Ideally, the utilization of the entire hyperspectral (HS) cube would have been preferred; however, due to the large size of the HS cube, memory limitations were encountered. To address this challenge, mitigating measures involved the reduction of spatial dimensions while preserving spectral characteristics. Consequently, the largest feasible patch size, specifically 128 by 128, was adopted, incorporating 214 spectral bands of the HS cube where the first ten bands were discarded due to the highly contained noise. It is noteworthy that while larger patch sizes could be accommodated by diminishing the batch size, such adjustments yielded suboptimal performance in our experiments. Specifically, experiments conducted with a batch size of 4 or below resulted in diminished model performance. The hyperparameters employed for this patch-based approach are outlined in table \ref{tab:Patches-hyper}.
\begin{table}
    \centering
    \caption{Patches training hyperparameters}
    \label{tab:Patches-hyper}
    \begin{tabular}{cc}
    \toprule
         Hyperparameter& Value\\
         \midrule
         Image resolution& 128x128x214\\
         Batch size& 8\\
         Optimizer& Adam\\
         Loss function& Weighted CCE\\
         Learning rate& 1e-5\\
 Class weights&.1 - .7 - .95 - .95\\
 Early stopping&20\\
 \bottomrule
    \end{tabular}
 
\end{table}

A comparative analysis of the hyperparameters utilized in the training scenarios for RGB and HSI PCA, as presented in tables \ref{tab:RGB-hyper} and \ref{tab:PCA-hyper}, along with table \ref{tab:Patches-hyper} reveals a notable distinction in the learning rate. Specifically, in the HSI patch training scenario, the learning rate is set at a lower value of 1e-5. This adjustment is not only empirically substantiated by observed results but is also theoretically justified due to the substantial increase in the number of training samples. The dataset size has significantly expanded from around a hundred instances in the PCA case and a couple of hundred in the RGB case to a couple of thousand instances in the patching scenario. For this resulting dataset, a smaller learning rate yielded better performance due to the smaller more repetitive parameter updates over more diverse training samples.

\subsubsection{\textbf{HSI Patches General Segmentation}}
Table \ref{tab:Patches-general} demonstrates the results of training three DL models on the 'General' ground truth. Three benchmark models were used instead of five due to incompatibilities between the hyperspectral data patches and DeepLabv3+ and LinkNet models' architecture. 
\begin{table}
    \caption{Models' evaluation metrics on the HSI patches test set 'General' ground truth.}
    \begin{adjustbox}{width=0.48\textwidth}
    \centering
    \begin{tabular}{ccccccc}
    \toprule
         \multicolumn{2}{c}{\textbf{Model}}&           \textbf{Metric $\backslash$ Class}&  \textbf{Others}&  \textbf{IC}   &  \textbf{Capacitor}   & \textbf{Connectors}\\ 
          \midrule 
         \multicolumn{2}{c}{\textbf{Unet}}&  Precision&  0.97&  0.76&  0.76& 0.39\\ 
         \multicolumn{2}{c}{}&  Recall   &  0.98&  0.58&  0.69& 0.70\\ 
         \multicolumn{2}{c}{}&  F1 Score &  0.98&  0.65&  0.72& 0.50\\ 
         \multicolumn{2}{c}{}&  IOU      &  0.96&  0.49&  0.57& 0.33\\ 
         \multicolumn{2}{c}{\textbf{Attention Unet}}&  Precision&  0.97&  0.75&  0.58& 0.18\\ 
         &  &  Recall   &  0.97&  0.55&  0.80& 0.79\\ 
         &  &  F1 Score &  0.97&  0.64&  0.37& 0.30\\ 
         &  &  IOU      &  0.95&  0.47&  0.51& 0.18\\ 
 \multicolumn{2}{c}{\textbf{ResUnet}}& Precision& 0.98& 0.72& 0.51&0.22\\ 
 & & Recall   & 0.97& 0.70& 0.86&0.70\\ 
 & & F1 Score & 0.98& 0.71& 0.64&0.34\\ 
 & & IOU      & 0.95& 0.55& 0.47&0.20\\ 
  \bottomrule
    \end{tabular}
    \end{adjustbox}
    \label{tab:Patches-general}
\end{table}
From table \ref{tab:Patches-general} it can be concluded that among the models, ResUnet and Unet demonstrate similar performance across multiple classes, with ResUnet having an edge in capturing only 'IC' whereas Unet was outperforming regarding the 'Capacitor' and 'Connctors' classes.

\begin{figure*}[t!]
  \centering
  \includegraphics[width=0.7\textwidth]{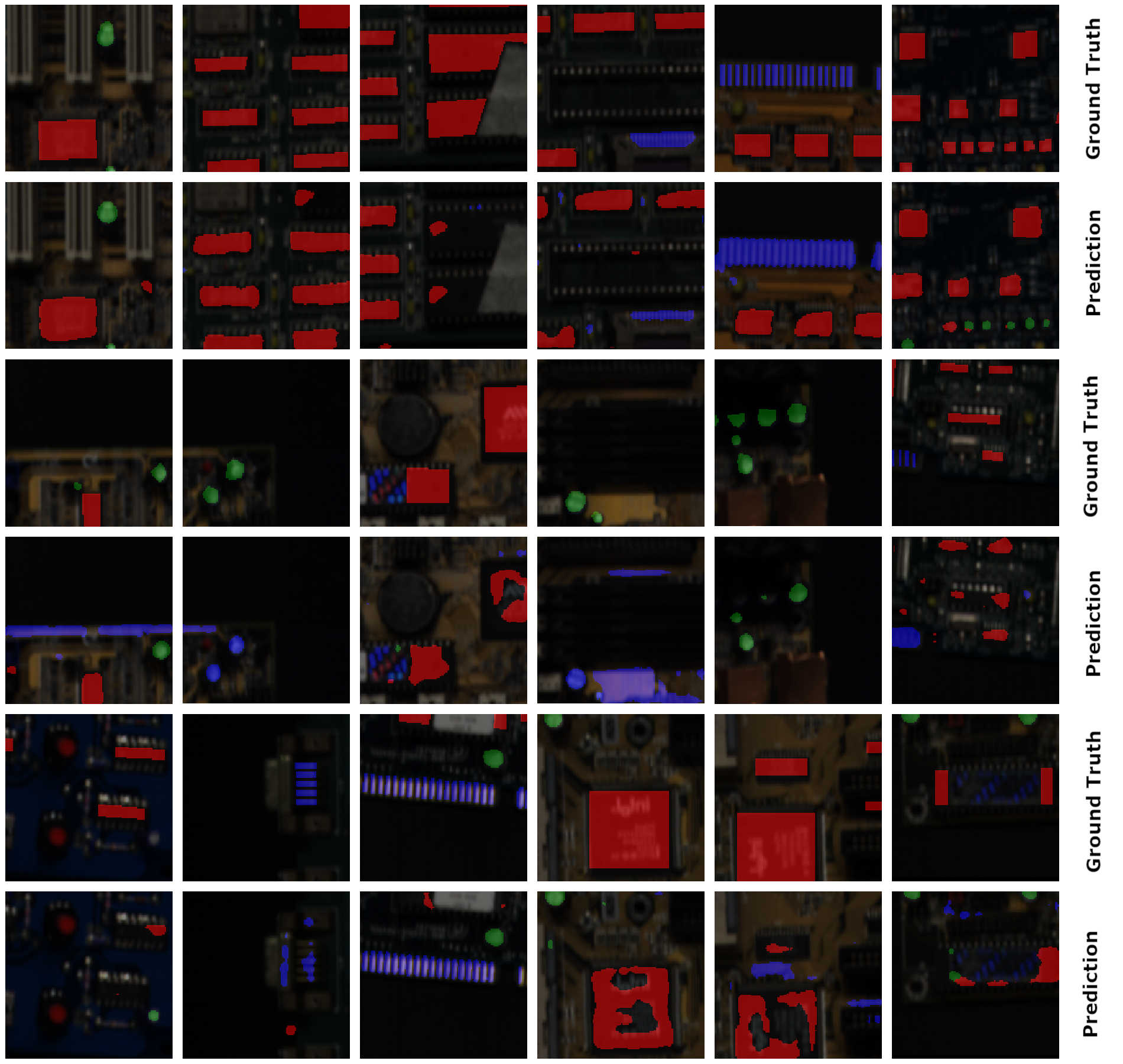}  
  \caption{Visual comparison of Unet predictions and ground truths on several HSI test patches.}
  \label{fig:Prediction Patches}
\end{figure*}

Figure \ref{fig:Prediction Patches} shows the segmentation prediction performance of U-Net on some test HSI patches of the three classes.

Unet performs well on the three classes, with a high overlap between the predicted and ground truth masks. However, there are some cases in which the model makes mistakes. Furthermore:
\begin{itemize}
    \item The fact that the model can achieve good segmentation performance on hyperspectral data which are much more information-rich and complex than RGB images is significant. This suggests that Unet has the potential to be used for a wide range of PCB HSI segmentation tasks.
    \item The model's performance on the 'Connectors' class is the weakest (examples in blue color), which suggests that there is room for improvement in this area. Future research could focus on developing new training techniques or architectures that can improve the model's performance on this class.
\end{itemize}

\subsubsection{\textbf{HSI Patches Monoseg Segmentation}}
Table \ref{tab:Patches-general} demonstrates the results of training the three DL models on the 'Monoseg' ground truth.
\begin{table}
\caption{Models' evaluation metrics on the HSI patches test set 'Monoseg' ground truth.}
\begin{adjustbox}{width=0.48\textwidth}
    \centering
    \begin{tabular}{ccccccc} 
    \toprule
         \multicolumn{2}{c}{\textbf{Model}}&           \textbf{Metric $\backslash$ Class}&  \textbf{Others}&  \textbf{IC}   &  \textbf{Capacitor}   & \textbf{Connectors}\\ 
         \midrule 
         \multicolumn{2}{c}{\textbf{Unet}}&  Precision&  0.97&  0.59&  0.73& 0.35\\ 
         \multicolumn{2}{c}{}&  Recall   &  0.98&  0.45&  0.64& 0.53\\ 
         \multicolumn{2}{c}{}&  F1 Score &  0.98&  0.51&  0.68& 0.42\\ 
         \multicolumn{2}{c}{}&  IOU      &  0.96&  0.34&  0.52& 0.27\\ 
         \multicolumn{2}{c}{\textbf{Attention Unet}}&  Precision&  0.98&  0.64&  0.65& 0.28\\ 
         &  &  Recall   &  0.98&  0.46&  0.73& 0.74\\ 
         &  &  F1 Score &  0.98&  0.54&  0.69& 0.41\\ 
         &  &  IOU      &  0.96&  0.37&  0.53& 0.26\\ 
 \multicolumn{2}{c}{\textbf{ResUnet}}& Precision& 0.98& 0.55& 0.51&0.18\\ 
 & & Recall   & 0.97& 0.64& 0.87&0.55\\ 
 & & F1 Score & 0.97& 0.59& 0.64&0.28\\ 
 & & IOU      & 0.95& 0.42& 0.47&0.16\\ 
  \bottomrule
    \end{tabular}
    \end{adjustbox}
    \label{tab:Patches-Monoseg}
\end{table}

Both Unet and Attention Unet perform consistently across different metrics and classes, showing high precision and recall, with an edge for the Unet model.

\section{\textbf{Discussion}}
In this section, we highlight the challenges that come along with performing segmentation on RGB and multiscene HS data cube.
\subsection{\textbf{RGB Segmentation}}
The segmentation performance across the RGB data reveals opportunities for further improvement. To enhance performance, increasing the dataset size beyond the current 53 (400 after augmentation) scenes is recommended as it is always the case in improving model generalization by providing more data, nevertheless, PCB-vision will be regularly updated with more scenes. Additionally, exploring more sophisticated deep learning models, leveraging pre-trained models, large vision models (LVMs) could contribute to improved segmentation results. The current benchmark models serve as a baseline, and further experimentation with larger datasets and advanced models is warranted.

\subsection{\textbf{HSI Segmentation}}
Segmenting multiscene hyperspectral data poses unique challenges due to the abundant information in hyperspectral cubes and lower spatial resolution compared to RGB images. Achieving high-performance segmentation requires models capable of extracting both spectral and spatial features efficiently. Several insights emerge from the HSI segmentation experiments:
\begin{itemize}
    \item Incompatibility of RGB models on HSI data: Directly applying RGB segmentation benchmark models to HSI data yields suboptimal results. The substantial increase in the number of channels (214 in HS compared to 3 in RGB) introduces more parameters in the input layers, implying more parameters to be optimized with the same PCB scenes, leading to harder challenges in achieving 
higher generalizing performance. 
    \item CNNs spatial/spectral feature extraction: CNNs excel in capturing spatial features as the above SOTA models proved. However, those models utilize only 2D CNNs and not 3D CNNs that incorporate spectral and spatial features simultaneously, yet, 3D CNNs are computationally demanding and slower than their 2D counterparts, making them unsuitable for industrial applications demanding fast real-time performance.
    \item Patching vs. PCA for memory efficiency: Utilizing patches of raw hyperspectral data addresses memory limitations but can disrupt spatial features, crucial for geometrically shaped objects like ICs and capacitors. Dimensionality reduction techniques like PCA provide an alternative, but the method of applying PCA worths consideration.
    \item  PCA implementation challenges: Solving the patching problem that solves the lack of memory problem can be done by reducing the dimension of the HS data using a dimensionality reduction technique like PCA. The principal components that dimensionality reduction techniques produce can capture more than 90\% of the variance in the data in only the first few components, making them suitable for the task. Usually, methods like PCA are implemented on the data cube first, then the result is given to train the classification model. This sequence utilizes the idea of extracting the spectral features using a dimensionality model first, then extracting the spatial features using a segmentation model second, combining the best of both models towards better segmentation performance. However, note that in the provided data (our application) we have multiple data cubes in the training set, validation set, and testing set. Therefore, two questions arise:
    \begin{enumerate}
        \item Should PCA be implemented from scratch on each HS cube in all training and testing sets, and the principal components will be then the segmentation model input? 
        \item Should all the training cubes be used to train one PCA, and then use that trained PCA to transform the dataset?
    \end{enumerate}

Implementing PCA on each hyperspectral cube independently introduces higher variance in results due to diverse compositions in different cubes. Thus posing a harder generalization challenge on the segmentation models.
Alternatively, training a single PCA on all cubes increases homogeneity but demands a PCA method compatible with large multiscene hyperspectral dataset. 

\item Impact of undesired background: Another point we would like again to highlight is the negative effect undesired background brings to the processing pipeline. Having an undesired background in the HSI skews the calculation across the processing pipeline:   
\begin{itemize}
    \item Dimensionality reduction impact: The presence of undesired pixels in hyperspectral data significantly influences principal component calculations, particularly when their quantity surpasses that of the desired pixels (i.e., classes of interest). This circumstance results in suboptimal principal components, leading to an imperfect representation of the target classes \cite{arbash2023masking}.
\item Segmentation challenges: Undesired background pixels introduce complexities during segmentation. The model encounters increased intricacy, particularly when undesired pixels exhibit noise or share similar spectra with one of the classes of interest (e.g., conveyor belt pixels having spectra akin to 'IC' pixels due to both being black polymers), the model experiences confusion in capsulizing the 'IC' class, potentially leading to suboptimal convergence \cite{arbash2023masking}.
\end{itemize}
To address these challenges, a comprehensive solution involving the application of background masking throughout the entire processing pipeline is yet to be implemented. By effectively masking out the undesired background pixels, the adverse effects on dimensionality reduction techniques and segmentation DL models can be mitigated, promoting more accurate and robust results.

\end{itemize}

\section{\textbf{Future Challenges}}
The above discussion highlighted empirical challenges that demand attention for achieving higher generalized performance in multiscene hyperspectral data analysis. Starting with data masking throughout the dataset, using the provided PCB masks in order to mitigate the effects of the undesired backgrounds. Furthermore, subsequent efforts involve the development of a multi-data type processing pipeline that seamlessly integrates RGB images with their hyperspectral pairs to overcome challenges unique to each data type. Additionally, we plan to continuously expand our PCB-Vision dataset by incorporating additional PCB scenes from diverse PCB sources and scanning sensors. This will help to better capture the variability in PCBs and improve the generalization ability of the models to unseen data.

The primary focus revolves around the refinement of segmentation models, with an emphasis on enhancing generalization performance within the processing pipeline to ensure optimal fast performance on unseen HS data. In this pursuit, additional research initiatives aim to provide pre-trained backbones tailored for segmentation and super-resolution models, expanding the scope of this work towards comprehensive advancements in hyperspectral data analysis.

\section{\textbf{Conclusion}}
In this work, we present a significant leap forward in PCB analysis along the generalized hyperspectral data processing concerning optimized E-waste recycling aligned towards a circular economy. We provide 'PCB-Vision', a pioneering RGB-HSI benchmark dataset of 53 different printed circuit boards (PCB), offering essential insights into E-waste composition, and setting the data basis for solution developments in sustainable product design and their recycling strategies.
Alongside the acquisition setup clarification and data description, we conducted an intensive statistical analysis of PCB-Vision. Moreover, we performed comprehensive experimentation with five SOTA segmentation models on both data types in PCB-Vision and highlighted the challenges and complexities inherent in classifying unseen hyperspectral data which is one of the main motivations for presenting the PCB-Vision benchmark dataset.
The reported results not only contribute valuable insights into the performance of benchmark models but also emphasize the need for models with robust generalization capabilities.
By addressing the detection of PCB elements towards object-based recycling, PCB-Vision supports the United Nations (UN) "Climate Action" SDG 13 and promotes innovation in electronic waste management, contributing to sustainable industrial practices outlined in SDG 9 "Industry, Innovation And Infrastructure".
As we look ahead, the outlined future challenges and proposed approaches aim to encourage further advancements in hyperspectral imaging and PCB analysis, fostering continued progress in this critical domain with the aim of optimized recycling.
To democratize scientific knowledge and promote inclusive innovation, we will openly release the benchmark dataset, ground truths, and accompanying baseline codes at \url{https://github.com/hifexplo/PCBVision}, enabling researchers from diverse fields to explore and contribute to the development of advanced E-waste recycling methodologies. This fosters resource collaboration and an open-access environment that aligns with the principles of the Green Deal.
This paper, therefore, represents a crucial step for achieving interconnected SDGs, fostering responsible consumption, reducing environmental impact, and advancing industrial sustainability.


\section{\textbf{Acknowledgment}}
The authors express their gratitude to EIT RawMaterials for funding the project 'RAMSES-4-CE' (KIC RM 19262). Appreciation is extended to the European Regional Development Fund (EFRE) and the Land of Saxony for their support in funding the computational equipment under the project 'CirculAIre.' Special thanks go to Junaidh Shaik Fareedh for his contributions to CAD designs and to Yuleika Carolina Madriz Diaz for her assistance in data acquisition.

\printbibliography




\end{document}